\documentclass{article}

\usepackage[final]{neurips_data_2024}

\usepackage[utf8]{inputenc} % allow utf-8 input
\usepackage[T1]{fontenc}    % use 8-bit T1 fonts
\usepackage{hyperref}       % hyperlinks
\usepackage{url}            % simple URL typesetting
\usepackage{booktabs}       % professional-quality tables
\usepackage{amsfonts}       % blackboard math symbols
\usepackage{nicefrac}       % compact symbols for 1/2, etc.
\usepackage{microtype}      % microtypography
\usepackage{xcolor}         % colors
\usepackage{graphicx}
\usepackage{amsmath}
\usepackage{subcaption}
\usepackage{colortbl}

\definecolor{iblue}{RGB}{0,0,156}
\definecolor{ired}{RGB}{192,0,0}
\definecolor{igrey}{RGB}{108,123,139}

\hypersetup{
	colorlinks=true,
	urlcolor=iblue,
	citecolor=iblue,
	linktoc=all,
	linkcolor=iblue
	}

\definecolor{iblue}{RGB}{0,0,156}
\definecolor{ired}{RGB}{192,0,0}
\definecolor{igrey}{RGB}{108,123,139}

\newcommand{\indep}{\perp \!\!\! \perp}

\usepackage{tikz}
\usetikzlibrary{shapes,decorations,arrows,calc,arrows.meta,fit,positioning}
\tikzset{
    -Latex,auto,node distance =1 cm and 1 cm,semithick ,
    state/.style ={ellipse, draw, minimum width = 0.7 cm},
    point/.style = {circle, draw, inner sep=0.04cm,fill,node contents={}},
    bidirected/.style={Latex-Latex,dashed},
    el/.style = {inner sep=2pt, align=left, sloped}
}

\title{FairJob: A Real-World Dataset \\
for Fairness in Online Systems}

\author{%
  Mariia Vladimirova\footnote{},\quad  Eustache Diemert \\
  Criteo AI Lab\\
  \texttt{\{m.vladimirova,e.diemert\}@criteo.com} 
  \And
  Federico Pavone \\
  Université Paris Dauphine-PSL \\
 \texttt{federico.pavone@dauphine.psl.eu} 
}

\begin{document}

\maketitle

\begin{abstract}
  We introduce a fairness-aware dataset for job recommendation in advertising, designed to foster research in algorithmic fairness within real-world scenarios. It was collected and prepared to comply with privacy standards and business confidentiality. An additional challenge is the lack of access to protected user attributes such as gender, 
  for which we propose a 
  solution 
  to obtain a proxy estimate. 
  Despite being anonymized and including a proxy for a sensitive attribute, our dataset preserves predictive power and maintains a realistic and challenging benchmark. This dataset addresses a significant gap in the availability of fairness-focused resources for high-impact domains like advertising -- the actual impact being having access or not to precious employment opportunities, where balancing fairness and utility is a common industrial challenge. We also explore various stages in the advertising process where unfairness can occur and introduce a method to compute a fair utility metric for the job recommendations in online systems case from a biased dataset. Experimental evaluations of bias mitigation techniques on the released dataset demonstrate potential improvements in fairness and the associated trade-offs with utility.

  The dataset is hosted at \href{https://huggingface.co/datasets/criteo/FairJob}{https://huggingface.co/datasets/criteo/FairJob}. 
 Source code for the experiments is hosted at \href{https://github.com/criteo-research/fairjob-dataset/}{https://github.com/criteo-research/FairJob-dataset/}. 
\end{abstract}

\section{Introduction}

The intersection of technology and human dynamics presents both opportunities and challenges, particularly in the realm of artificial intelligence (AI). Despite advancements, persistent biases rooted in historical inequalities permeate our data-driven systems, perpetuating unfairness and exacerbating societal divides. 
Historical biases shape data collection, influencing AI model outcomes and often  \textit{amplifying} existing inequalities~\citep{bolukbasi2016man, zhao2017men,chen2023bias}.
Despite concerns regarding privacy, liability, and public relations, the collection of special and sensitive category data is crucial for bias assessments~\citep{andrus2021we}. Moreover, evolving legal frameworks, exemplified by the recent AI Act and General Data Protector Regulation~\citep{uk2022gdpr}, mandate the detection, prevention, and mitigation of biases, while imposing some restrictions on the use of sensitive data.

Recent advances in fairness often involve computer vision, natural language processing and speech recognition tasks~\citep{gustafson2023facet,andrews2024ethical,hall2024visogender,schumann2024consensus,veliche2023improving}, while lacking attention to algorithmic decision-making that involves \textit{tabular data}, where each row represents an individual or an observation, and each column represents a feature or attribute~\citep{le2022survey,zhang2021ai}, resulting in a very few benchmark papers~\citep{Gorishniy2021Revisiting,gorishniy2022embeddings,Grinsztajn2022why,shwartz2022tabular, matteucci2023benchmark}. Tabular data is commonly used in various \textit{high-risk domains} such as finance, healthcare, hiring, criminal justice, and advertising~\citep{van2024tabular}. 

Algorithmic discrimination in advertising can be related to sensitive verticals which highlights beneficial employment, financial and housing opportunities, or about who sees potentially less desirable advertising, such as ads for predatory lending services~\citep{lambrecht2019algorithmic}. While unfairness in advertising is not punitive but rather assistive, i.e. fairness consists in providing equal access to precious opportunities, it is essential to \textit{ensure fairness in advertising practices}. In some contexts such as housing or lending, such discrimination is explicitly \textit{prohibited by law}\footnote{According to Article 6(2) of the AI Act, targeted job advertising is considered as high-risk AI system. The Fair Housing Act in the United States makes it illegal to discriminate based on religion, color, national origin or gender for the sale, rental or financing of housing.}. 
% Thus, it necessitates proactive measures to detect, prevent, and mitigate biases.
Several studies conducted analyses on the fairness in advertising at different stages and observed discriminating behavior that was not necessarily intended by the ad-services~\citep{speicher2018potential,lambrecht2019algorithmic,andreou2019measuring,ali2019discrimination}. 
This emphasizes the need for
better mechanisms to audit and prevent bias in ads.

Most of studies on discriminating behavior in advertising were conducted via creating advertising campaigns and choosing targeted audiences and analysing the data from the user perspective without accessing the algorithmic features~\citep{speicher2018potential,lambrecht2019algorithmic,andreou2019measuring,ali2019discrimination}.  
The absence of publicly available, realistic datasets leads researchers to publish results based on private data, resulting in non-reproducible claims~\citep{geyik2019fairness,andreou2019measuring,timmaraju2023towards,tang2022towards}. This poses challenges for critical evaluation and building upon previous work in the scientific community. \citet{tang2022towards} highlights the lack of public benchmarking datasets to study the fairness related approaches in advertising.

In addition, most of the studies assume that the AI systems have an access to the protected attributes which is often \textit{unrealistic} due to privacy constraints or legal restrictions~\citep{holstein2019improving,lahoti2020fairness,molina2023trading,timmaraju2023towards}. In online advertising, decision-makers usually have access to a log of user interactions with the system, which they can use to guess the attributes. However, the level of inaccuracy can be significant, making it difficult to ensure that an ad campaign reaches a non-discriminatory audience~\citep{gelauff2020advertising}. This makes it hard to meet fairness requirements~\citep{lipton2018does}. We emphasize the need for thorough research in real-world situations where \textit{access to protected attributes is limited}.

\paragraph{Contributions.} To foster research in fairness within real-world scenarios, we release a large-scale fairness-aware dataset for advertising. The dataset contains pseudononymized users' context and publisher features that were collected from a job targeting campaign ran for 5 months. The data has been \textit{sub-sampled non-uniformly} to avoid disclosing business metrics. Feature names have been \textit{anonymized} for business confidentiality, and their \textit{values randomly projected} to preserve predictive power while making the \textit{recovery of the original features or user context (i.e. re-indentification) practically impossible}, with accordance to the privacy-safety measures\footnote{'Pseudonymisation' of data -- defined in Article 4(5) of General Data Protection Regulation (GDPR) -- means replacing any information which could be used to identify an individual with a pseudonym, or, in other words, a value which does not allow the individual to be directly identified. The orginal data is still used in the AdTech company, however, re-indentification of the pseudomymized data is impossible due to additional randomization techniques during data anonymization. The original data does not contain any "special categories" of personal data listed under Article 9 of the GDPR, processing of which is prohibited, except in limited circumstances set out in Article 9 of the GDPR. The publication of the dataset was approved by a Data Protector Officer and Legal professionals.}. Although our dataset does not contain explicit sensitive attributes such as gender, it includes a \textit{gender proxy derived from non-protected relevant attributes}, which we discuss in detail further.

This dataset provides a baseline according to the eligible audience generated by an advertiser's targeting criteria for a specific ad. This ensures that ads are tailored to individuals whom the advertiser can feasibly serve (such as those within a specific geographic region) and who are likely to be interested in their offerings, a practice already \textit{governed by policies and standards in Housing, Employment, and Credit verticals}. Since advertiser targeting \textit{adheres to policy constraints to prevent discriminatory practices}\footnote{The Fair Housing Act in the United States makes it illegal to discriminate based on religion, color, national origin or gender for the sale, rental or financing of housing.} (such as prohibiting the use of gender criteria in employment ads), the resulting eligible audience remains independent of prediction algorithms, serving as a reasonable baseline metric. 

With the released dataset we examine the stages in the advertising process where unfairness can occur and explore techniques to mitigate such biases. Taking into account possible induced biases, we propose an unbiased utility metric that help to analyse different bias mitigation techniques. We also perform experiments on the released dataset to verify how we can improve fairness and the possible trade-offs with utility.

\section{Related works}

\paragraph{Open-source datasets.} 
A limited availability of publicly available fairness-aware tabular datasets challenges research advancements in algorithmic fairness~\citep{le2022survey,hort2023bias}. 
In 2022, \citet{le2022survey} studied datasets used at least 3 times in research publications on fairness, and found out there were only 15 open-source fairness datasets, most of which are criticized for being too small or far from real-world scenarios, including the most frequently used Adult~\citep{dua2017adult} and COMPAS dataset~\citep{larson2016compas}. Even though there is a positive tendency on addressing this issue by open-sourcing privacy-complying datasets, such as BAF~\citep{jesus2022turning} for bank fraud detection where the data was obtained via data generation techniques, or WCLD~\citep{ash2024wcld}, a curated large-scale dataset from circuit courts to address criminal justice, there is still \textit{lack in available datasets} in other high-impact areas such advertising.
It is important for academic researchers to have access to large datasets to study the problem rigorously~\citep{cardoso2019framework,li2022data,le2022survey}. Large-scale datasets are advantageous as they increase the likelihood of capturing significant performance differences in experiments with new methods. With larger dataset sizes, the variance of metrics decreases, enabling more reliable and meaningful comparisons between different approaches.

\paragraph{Bias mitigation methods.}
The initial step to enhance model fairness is to exclude the protected attribute as a feature during training, a strategy known as \textit{fairness through unawareness}~\citep{chen2019fairness}. 
However, this approach alone does not ensure fairness because the model may still learn correlations between other features and the protected attributes, see Section~\ref{subsec:fairness_definition} and Figure~\ref{fig:second} for details. To achieve a higher level of fairness, AI systems typically employ one of the additional methods: \textit{pre-processing, in-training, or post-processing}. We refer to~\citet{hort2023bias} for the most up-to-date and thorough survey.

\paragraph{Fairness without demographics.} The information on the protected attribute is often not available in practice~\citep{holstein2019improving,hort2023bias}. 
Several works studied limited availability of the protected attribute such as via a proxy~\citep{gupta2018proxy} or assuming there is a partial access to the information~\citep{hashimoto2018fairness,awasthi2020equalized,molina2023trading}. 
\citet{lahoti2020fairness} relies on the assumption that protected groups are computationally-identifiable.
However, if there were no signal about protected groups in the remaining features and class labels, we cannot make any statements about improving the model for protected groups. 
One of the possible solutions is to get data from secure multi-party computation~\citep{veale2017fairer,kilbertus2018blind,hu2019distributed} or directly from users~\citep{gkiouzepi2023collaborative}. However, these tools are still to be adapted to real-world situations. In addition, transfer leaning can be useful when there is little available data on the protected attributes~\citep{coston2019fair}.

\section{Fairness in advertising}
\label{sec:fairness_in_advertising}

The aim of ad-tech companies is to deliver the most relevant advertisements to users navigating publishers' webpages. By matching users' browsing histories and content preferences with products that align with their interests, targeted advertising creates a mutually beneficial ecosystem~\citep{wang2017display,choi2020online}. Advertisers reach relevant audiences, users have access to free information and services in exchange of seeing ads related to their interests, and platforms profit from selling targeted ads. 

Ad-tech companies grapple with vast volumes of noisy data, which encapsulate users' past actions. Leveraging this data, they predict potential clicks and conversions. However, if the data is biased, the algorithms can inadvertently perpetuate and even amplify these biases~\citep{bolukbasi2016man, zhao2017men,chen2023bias}. It is crucial to scrutinize the predictors for bias and devise solutions to mitigate it. 
Failing to do so can result in discrepancies between offline evaluations and online metrics, ultimately harming user satisfaction and trust in the service of online systems~\citep{chen2023bias}. While advertising commonplace items carries little risk, companies must exercise caution with high-risk verticals like job offers~\citep{speicher2018potential,lambrecht2019algorithmic,andreou2019measuring,ali2019discrimination}. For instance, \textit{if managerial positions are disproportionately shown to men over women, more men may apply, perpetuating historical biases and exacerbating gender disparities}.

Bias can be introduced at several stages in the advertising process, see Figure~\ref{fid:advertising_process}. First, when a user visits a webpage with an ad slot, ad-tech companies participate in a real-time bidding (RTB) auction. During this auction, companies select a campaign (e.g., job offers or clothing) based on attributes of the publisher and the user, including their log of past interactions such as seen ads, their context, the fact of clicks on the ads, see Section~\ref{subsection:selection_bias_campaign}. This auction must be organized in a fair way, respecting both the companies placing bids and the publishers providing ad slots, see Section~\ref{subsec:market_unfairness}.
After an ad-tech company wins the display auction, there is the choice of which product to show (e.g., a senior position job or an assistant job). This selection can also introduce bias with respect to the user, see Section~\ref{subsec:recommendation_bias}. Ensuring fairness at this stage is critical to preventing the reinforcement of existing inequalities.

\definecolor{34f7703b-c631-5d19-a2aa-8444d08d9619}{RGB}{255, 255, 255}
\definecolor{f3551e38-74df-57e2-b793-83d7fe876c85}{RGB}{0, 0, 0}
\definecolor{0b71a967-1f15-55a5-9bb9-70efa7b4fc58}{RGB}{51, 51, 51}

\tikzstyle{ab01b2d5-7def-578f-a73f-b27ff04e136c} = [rectangle, rounded corners, minimum width=2cm, minimum height=1cm, text centered, font=\normalsize, color=0b71a967-1f15-55a5-9bb9-70efa7b4fc58, draw=f3551e38-74df-57e2-b793-83d7fe876c85, line width=1, fill=34f7703b-c631-5d19-a2aa-8444d08d9619]
\tikzstyle{ad4eee00-ec60-54a2-8406-2c673a76cc69} = [thick, draw=f3551e38-74df-57e2-b793-83d7fe876c85, line width=1, ->, >=stealth]

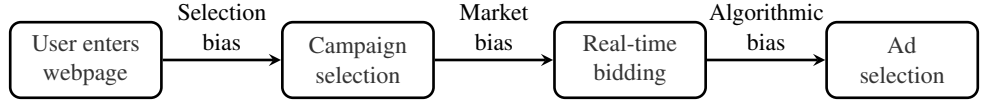
\begin{figure}[ht]
\centering
\begin{tikzpicture}[node distance=2cm]

\node (a3298b4f-b680-4781-90be-d6dbb35896b2) [ab01b2d5-7def-578f-a73f-b27ff04e136c] {\begin{minipage}[lt]{42.19pt}\setlength\topsep{0pt}
\begin{center}
{\footnotesize User enters }\\{\footnotesize webpage}
\end{center}

\end{minipage}};
\node (71ec9b60-f9bb-4068-8d26-866144ab80a0) [ab01b2d5-7def-578f-a73f-b27ff04e136c, right of=a3298b4f-b680-4781-90be-d6dbb35896b2, xshift=1.6cm] {\begin{minipage}[lt]{42.19pt}\setlength\topsep{0pt}
\begin{center}
{\footnotesize Campaign }\\{\footnotesize selection}
\end{center}

\end{minipage}};
\node (42830576-d71a-4308-95f9-790be3d0ae10) [ab01b2d5-7def-578f-a73f-b27ff04e136c, right of=71ec9b60-f9bb-4068-8d26-866144ab80a0, xshift=1.6cm] {\begin{minipage}[lt]{42.19pt}\setlength\topsep{0pt}
\begin{center}
{\footnotesize Real-time }\\{\footnotesize bidding}
\end{center}

\end{minipage}};
\node (0bc33426-a08d-4185-88a5-69936e9fc020) [ab01b2d5-7def-578f-a73f-b27ff04e136c, right of=42830576-d71a-4308-95f9-790be3d0ae10, xshift=1.6cm] {\begin{minipage}[lt]{42.19pt}\setlength\topsep{0pt}
\begin{center}
{\footnotesize Ad }\\{\footnotesize selection}
\end{center}

\end{minipage}};
\draw [ad4eee00-ec60-54a2-8406-2c673a76cc69] (a3298b4f-b680-4781-90be-d6dbb35896b2) -- node  {\begin{minipage}[lt]{42.19pt}\setlength\topsep{0pt}
\begin{center}
{\footnotesize Selection }\\{\footnotesize bias}
\end{center}
\end{minipage}} (71ec9b60-f9bb-4068-8d26-866144ab80a0);
\draw [ad4eee00-ec60-54a2-8406-2c673a76cc69] (71ec9b60-f9bb-4068-8d26-866144ab80a0) -- node  {\begin{minipage}[lt]{42.19pt}\setlength\topsep{0pt}
\begin{center}
{\footnotesize Market }\\{\footnotesize bias}
\end{center}
\end{minipage}} (42830576-d71a-4308-95f9-790be3d0ae10);
\draw [ad4eee00-ec60-54a2-8406-2c673a76cc69] (42830576-d71a-4308-95f9-790be3d0ae10) -- node  {\begin{minipage}[lt]{42.19pt}\setlength\topsep{0pt}
\begin{center}
{\footnotesize Algorithmic }\\{\footnotesize bias}
\end{center}
\end{minipage}} (0bc33426-a08d-4185-88a5-69936e9fc020);
\end{tikzpicture}
\caption{Simplified scheme of online advertising process of ad selection: (i) user enters a webpage with available banner for an ad, (ii) webpage sends a request to participate in the real-time bidding auction which triggers campaign selection by an ad service for a given user, (iii) after the campaign is chosen, ad-service sends a bid proposition, (iv) if the proposed bid won the auction, the recommendation engine chooses the best ad from the chosen campaign and shows it on the webpage.}
\label{fid:advertising_process}
\end{figure}

\subsection{Fairness definition}
\label{subsec:fairness_definition}

We base our discussion on a counterfactual fairness framework that explains the underlying connections between the variables in the system~\citep{kusner2017counterfactual}. Let $A$ denote a protected attribute (can be a set of protected attributes) of an individual, $X$ denote the other observable attributes of any particular individual, $Y$ denote the outcome to be predicted, and let $\hat{Y}$ be a predictor. The predictor takes into account the available data from logs of user interactions with the system and product descriptions and estimates the probability of a positive outcome, i.e. click of the user on the product. The system takes into account the prediction and then shows the best product to the user, which results into a possible positive outcome.
In our analysis, we are interested in understanding how $A$ and $X$ influence $Y$ and how well our predictor $\hat{Y}$ captures these relationships. Our goal is not just to predict outcomes accurately but also to ensure fairness and mitigate biases in the predictions with respect to $A$.
These random variables exhibit causal relationships, as modeled in Fig.~\ref{fig:causal_graph}, which we further explore in detail below.

\tikzstyle{nodes} = [circle, minimum width=1cm, minimum height=1cm, text centered, font=\normalsize, color=0b71a967-1f15-55a5-9bb9-70efa7b4fc58, draw=f3551e38-74df-57e2-b793-83d7fe876c85, line width=1, fill=34f7703b-c631-5d19-a2aa-8444d08d9619]

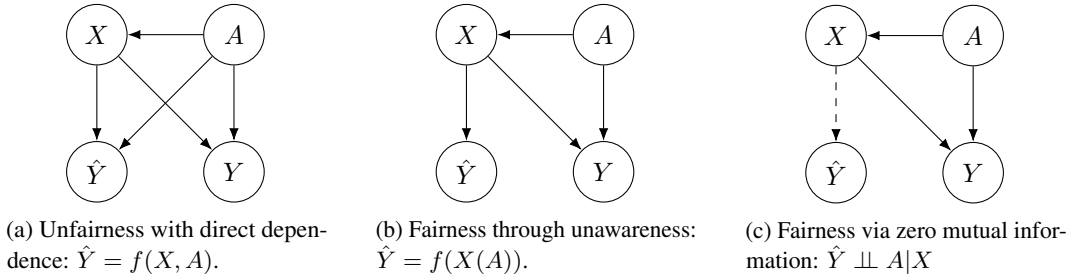
\begin{figure}[ht]
\centering
\begin{subfigure}{0.3\textwidth}
\centering
    \begin{tikzpicture}[scale=0.3]
        \node[state, circle, minimum width=0.8cm] (x) {$X$};
        \node[state, circle, minimum width=0.8cm] (a) [right =of x]  {$A$}; 
        \node[state, circle, minimum width=0.8cm] (haty) [below =of x] {$\hat{Y}$};
        \node[state, circle, minimum width=0.8cm] (y) [right =of haty] {$Y$};
        
        \path (x) edge (haty);
        \path (a) edge (y);
        \path (a) edge (x);
        \path (a) edge (haty);
        \path (x) edge (y);
    \end{tikzpicture}
    \caption{Unfairness with direct dependence: $\hat{Y} = f(X, A)$.}
    \label{fig:first}
\end{subfigure}
\hfill
\begin{subfigure}{0.3\textwidth}
\centering
    \begin{tikzpicture}[scale=0.33]
        \node[state, circle, minimum width=0.8cm] (x) {$X$};
        \node[state, circle, minimum width=0.8cm] (a) [right =of x]  {$A$}; 
        \node[state, circle, minimum width=0.8cm] (haty) [below =of x] {$\hat{Y}$};
        \node[state, circle, minimum width=0.8cm] (y) [right =of haty] {$Y$};
        
        \path (x) edge (haty);
        \path (a) edge (y);
        \path (a) edge (x);
        \path (x) edge (y);
    \end{tikzpicture}
    \caption{Fairness through unawareness: $\hat{Y} = f(X(A))$.}
    \label{fig:second}
\end{subfigure}
\hfill
\begin{subfigure}{0.3\textwidth}
\centering
    \begin{tikzpicture}[scale=0.33]
        \node[state, circle, minimum width=0.8cm] (x) {$X$};
        \node[state, circle, minimum width=0.8cm] (a) [right =of x]  {$A$}; 
        \node[state, circle, minimum width=0.8cm] (haty) [below =of x] {$\hat{Y}$};
        \node[state, circle, minimum width=0.8cm] (y) [right =of haty] {$Y$};
        
        \path [draw, dashed] (x) edge (haty);
        % \path (haty) edge (y);
        \path (a) edge (y);
        \path (a) edge (x);
        \path (x) edge (y);
    \end{tikzpicture}
    \caption{Fairness via zero mutual information: $\hat{Y} \indep A | X$}
    \label{fig:third}
\end{subfigure}
         \caption{Causal graph depicting effects of variables appearing during model training under different constraints. The arrow between the nodes corresponds to the causal effect. The dashed arrow between $\hat{Y}$ and $X$ can be interpreted as $\hat{Y}$ depends on $X$, but conditionally on $X$, $\hat{Y}$  is independent to its ancestors.}
         \label{fig:causal_graph}
\end{figure}

It is important to understand how the information flows during the training procedure to create the prediction. If we give a protected attribute as a feature to the prediction model during training like in Fig.~\ref{fig:first}, the model will learn directly the bias $A \to \hat{Y}$. The first step is to remove the protected attribute from the training features, in this case the model might learn the bias indirectly through the features $A \to X \to \hat{Y}$, see Fig.~\ref{fig:second}. Thus, without bias mitigation techniques, the prediction model learns the bias that exists in the data. From causal perspective, the correction techniques correspond to blocking the information flow from $A$ to $\hat{Y}$ by enforcing the zero mutual information between these variables conditioned on $X$, see Fig.~\ref{fig:third}. We refer to  \citet{hort2023bias} for the survey on bias mitigation methods. Theoretically, we can mitigate the protected attribute bias when having access to the information that is used by an algorithm during training and having at least partial information about the protected attribute~\citep{lahoti2020fairness,hort2023bias}.

From a causal perspective, fair outcome with respect to a protected attribute means that it would not have changed if the other (counterfactual) value for the protected attributed was imposed~\citep{kusner2017counterfactual}:
\begin{equation}
    \mathbb{P} \left(Y = y\, \Bigl| \, do(A = a), X = x \right) = \mathbb{P} \left(Y = y\, \Bigl| \, do(A = a'), X = x \right).
\end{equation}
Since in a real-world scenarios it is not possible to have counterfactual estimation (we cannot impose a user to change their gender), we consider average values for groups, e.g. demographic parity\footnote{We do not use \textit{equal opportunity} metric because despite the name it actually preserves exiting bias in the case of assistive fairness, see Appendix for details.}: 
\begin{equation}
\label{eq:demographic_parity}
    \mathbb{P} \left( Y = y | A = a\right) = \mathbb{P} \left( Y = y | A = a'\right)
\end{equation}
This is in line with current laws that aim to ensure fairness in how housing and job ads are presented\footnote{The Fair Housing Act in the United States makes it illegal to discriminate based on religion, color, national origin or gender for the sale, rental or financing of housing.}. These laws do not focus on whether certain people are wrongly included or excluded (individual fairness), rather on making sure the ads are representative (group fairness). The key measurement is the difference in average that different groups will be shown the ad, regardless of how likely each group is to actually respond to the ad.

\subsection{Selection bias in campaign choosing}
\label{subsection:selection_bias_campaign}
In our setting, we are interested in assessing the fairness in specific campaign (e.g., job campaign) with respect to the protected attribute. For instance, we want to ensure that job advertisements for managerial roles are fair with respect to a binary protected attribute $A\in\{0,1\}$ (e.g., gender).
Typically, the data considered in this framework regards the job advertisements for users which have been assigned to the job campaign $c$. However, the campaign selection process might introduce selection bias, which should be taken in account. 
In particular, $\mathbb{P}(A=1)$ and $\mathbb{P}(A=0)$ are the (internet)-population level of a binary protected attribute. This might be approximated to the census population frequencies of the protected attribute. Let $C$ be a random variable of choosing a campaign, then $\mathbb{P}(A=1\mid C=c)$ and $\mathbb{P}(A=0\mid C=c)$ are the frequencies of the protected attribute in the job campaign data $c$. These differ from the population levels due to selection bias.

Note that the recommendation engines predict $\mathbb{P}(Y = 1\mid A=a, C=c)$ for a product in the campaign $c$. Thus, if we use prediction bias mitigation techniques while considering data at the campaign level, in the best case scenario, we obtain fair predictions while being unfair outside of campaing, $\mathbb{P}(\hat{Y} = 1\mid A=0, C=c) = \mathbb{P}(\hat{Y} = 1\mid A=1, C=c)$ and $\mathbb{P}(\hat{Y} = 1\mid A=0) \not= \mathbb{P}(\hat{Y} = 1\mid A=1)$ . Thus, we have to take into account the selection bias to ensure demographic parity introduced in Eq.~\eqref{eq:demographic_parity}. Details on the derivation of the campaign selection bias and its correction are referred to supplemental material. 

\subsection{Market bias}
\label{subsec:market_unfairness}
\citet{lambrecht2019algorithmic} found that women are a prized demographic, making them more expensive to
advertise to. This implies that ads that are meant to be gender-neutral can be delivered in the way
that appears to be discriminatory by RTB algorithms that focus on optimizing cost-effectiveness.
\citet{ali2019discrimination} explained that this is not solely the indication of the ingrained cultural bias nor a
result of user profiles inputted into ads algorithms, but rather the product of competitive spillovers
among advertisers. 
Additionally, the feedback loop mechanism considers imbalanced information--how recommendation systems expose content influences user behavior, which then becomes the training data for future predictions. This feedback loop not only introduces biases but also amplifies them over time, leading to a 'rich get richer' scenario known as the Matthew effect.~\citep{chen2023bias}.
Imbalanced data with respect to a protected attribute also effects the learning of a prediction, since an algorithm that receives in real time less data about one group, will learn at different speeds~\citep{lambrecht2020apparent}. 
These effects are hard to estimate and should be addressed by the RTB process. Apart from users, advertisers can also be unfairly treated during the RBT auction process~\citep{celis2019toward,chen2023bias} but here we focus solely on the user discrimination.

\subsection{Recommendation bias}
\label{subsec:recommendation_bias}

\tikzstyle{nodes} = [circle, minimum width=1cm, minimum height=1cm, text centered, font=\normalsize, color=0b71a967-1f15-55a5-9bb9-70efa7b4fc58, draw=f3551e38-74df-57e2-b793-83d7fe876c85, line width=1, fill=34f7703b-c631-5d19-a2aa-8444d08d9619]

\begin{figure}[ht]
\centering
\begin{subfigure}{0.49\textwidth}
\centering
    \begin{tikzpicture}[scale=0.3]
        \node[state, circle, minimum width=0.8cm] (x) {$X$};
        \node[state, circle, minimum width=0.8cm] (a) [right =of x]  {$A$}; 
        \node[state, circle, minimum width=0.8cm] (d) [below =of x] {$D$};
        \node[state, circle, minimum width=0.8cm] (haty) [right =of d] {$\hat{Y}$};
        
        \path (x) edge (haty);
        \path (d) edge (haty);
        \path (a) edge (x);
        \path (x) edge (d);
    \end{tikzpicture}
    \caption{Indirect unfairness: $\hat{Y} = f(X(A), D(A))$.}
    \label{fig:reco_first}
\end{subfigure}
\hfill
\begin{subfigure}{0.49\textwidth}
\centering
    \begin{tikzpicture}[scale=0.33]
        \node[state, circle, minimum width=0.8cm] (x) {$X$};
        \node[state, circle, minimum width=0.8cm] (a) [right =of x]  {$A$}; 
        \node[state, circle, minimum width=0.8cm] (d) [below =of x] {$D$};
        \node[state, circle, minimum width=0.8cm] (haty) [right =of d] {$\hat{Y}$};
        
        \path (x) edge (haty);
        \path (d) edge (haty);
        \path (a) [draw, dashed] edge (x);
        \path (x) edge (d);
    \end{tikzpicture}
    \caption{Fairness, zero mutual information: $\hat{Y} \indep A | X, D$}
    \label{fig:reco_second}
\end{subfigure}
         \caption{Causal graph depicting effects of variables appearing during model training for an ad recommendation system under different constraints.}
         \label{fig:reco_causal_graph}
\end{figure}
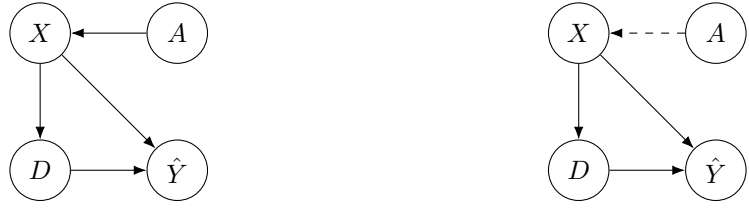

In the ad recommendation system, the goal is to choose best products for a user for a given banner that can have several displays at the same time. The goal is to maximize the number of clicks for a given banner, meaning that there can be several products clicked. When we have several displays to show to a user, the display rank position becomes important and creates position bias with respect to a positive outcome. The influence of this bias is hard to estimate, however, it is important to take it into account~\citep{singh2018fairness,singh2019policy,morik2020controlling,usunier2022fast}. 

Let $J$ be a random variable denoting the set of banner to be shown to a user, $D$ be a display (chosen product, i.e. job offer) shown to a user on a banner $J$. Let model $f(x, d)$ predicts the following positive outcome: $\mathbb{P}(Y = 1 | X = x, D = d)$, i.e. the probability of a click for a chosen product $d$ given user features $x$. As discussed above, we have to take into account the display position which expressed via variable rank $R$. However, the influence of the position on the utility is hard to estimate. Further, we suggest utility metrics for ads recommendation and in order to avoid the position bias, we suggest to compute them only on randomized displays, where the position of the products on the banner was chosen randomly. 

\paragraph{Click-rank utility.}
The users' utility for a given model can be expressed as a positive engagement in the following way: 
\begin{equation}
    U(f) =\mathbb{E}_{J} \mathbb{E}_{X, D\mid J} \bigl[\mathbb{I} \left(Y_{D} = 1 \right) \,  \text{rank}_{J} f(X_J, D)\bigr],
\end{equation}

where $\mathbb{I}(Y_D=1)$ is the identity function of a positive outcome (e.g. click) for display $D$. The function $\text{rank}_{D}$ computes the ascending order rank within the set of displays for a banner $J$. This metrics is based on estimation of the positive outcome based on the passed events for chosen users. 

\paragraph{Product-rank utility for biased data.}
We notice that the metrics for the algorithm can be biased due to the selection bias discussed in Section~\ref{subsection:selection_bias_campaign} because  
the prediction algorithm estimates $\mathbb{P}(Y = 1 | A = a, C = c)$ instead of $\mathbb{P}(Y = 1 | A = a)$. Even if we correct the prediction bias in $\mathbb{P}(Y = 1 | A = a, C = c)$ based on the data provided for given campaign $c$, it does not correct the final bias in $\mathbb{P}(Y = 1 | A = a)$ due to selection bias.  
We can adapt the click-rank utility to include possible selection bias into the metric, by explicitly considering that the product utility depends on a chosen campaign. Then, when correcting for the unfairness in the prediction, we might improve the utility metric taken into account the selection bias in the data:
\begin{equation}
    \tilde U(f) =\mathbb{E}_{D} \mathbb{E}_{J\mid D} \left[\mathbb{I} \left(Y_{D} = 1 \right) \, \frac{\mathbb{P}(A = a_{X_J})}{\mathbb{P}(A = a_{X_J} | C = c)} \, \text{rank}_J f(X_J, D)  \right],
\end{equation}
where $a_X$ stands for a gender of a given user $X$. 
Intuitively, if the prediction is biased with respect to the protected attribute $A$, the final prediction $\mathbb{P}(Y = 1 | A = a)$ is even more biased due to selection bias with respect to the protected attribute of choosing a campaign $C = c$: $\mathbb{P}(C=c\mid A=a)$. In this case, the prediction model amplifies the existing historical bias. However, we can remove the selection bias by adding weights that correspond to the presence of the protected attribute in the whole population and given the campaign. If the user with protected attribute $A = a$ has lower probability of click, and this group was underrepresented in the campaign $C = c$, i.e. $\mathbb{P}(A = a) > \mathbb{P}(A = a | C = c)$, then in the utility function, the model's prediction will be higher, by addressing the possible bias due to under-representation in the data. This is our suggested metric to evaluate the recommendation system when the selection bias is present and known such as in the FairJob dataset. 

\section{FairJobs dataset}

We introduce FairJobs\footnote{https://huggingface.co/datasets/criteo/FairJob} dataset that contains fairness-aware data from a real-world scenario of advertising. The intended use of this dataset is to learn click predictions models and evaluate by how much their predictions are biased between different gender groups. 
The dataset consists of 1,072,226 rows that were collected during 5 months of a targeted job campaign\footnote{We leave the details on the data collection, feature engineering and privacy-preserving steps in the supplemental material.}, each row represents a job ad and user features: 20 categorical and 39 numerical features; label \texttt{click} (binary, if the ad was clicked), \texttt{protected\_attribute} (binary, proxy for user gender, see below for more thorough explanation), \texttt{senior} (binary, if the job offer was for a senior position), [\texttt{user\_id}, \texttt{impression\_id}, \texttt{product\_id}] are unqiue identifiers of user, impression and product (job ad). More details and dataset statistics are referred to Appendix.

\paragraph{Details on gender proxy.}

Since we do not directly access user demographics, we have to find a way to get a proxy of relevant attribute\footnote{This gender proxy is used with the AdTech company to create in-market audiences: when a client can choose to show advertising of their products to “people that buy more female products” or “people that buy more male products”. We note that \textit{this proxy is not used in the prediction engine} and for campaign creation \textit{on high-risk verticals}. The campaign creation is governed by policies and standards in Housing, Employment, and Credit verticals that complies with the Fair Housing Act in the USA.}. Most of recent works leverage the use of external data or prior knowledge on correlations to obtain proxies to relevant attributes~\citep{gupta2018proxy,hashimoto2018fairness,awasthi2020equalized,lahoti2020fairness}. 
We define a product gender, either given by a client, either by a category of the product. This gives us approximately 40\% of products gender identified. Then, we follow the available statistics and choose the gender proxy based on the dominant gender of products the user interacts with. This gender proxy identities a behavior of a user, i.e. if a user tends to buy female or male products. The gender proxy does not necessarily correlate with the gender, as it often happens with the proxy variables~\citep{gelauff2020advertising}. Verification of the accuracy of these approximations is challenging. Additionally, if there are no signal about protected groups in the remaining features and class labels, we cannot make any statements about improving the model for protected groups~\citep{lahoti2020fairness}.

\paragraph{Limitations and interpretation.}  We remark that the proposed gender proxy does not give a definition of the gender. Since we do not have access to the sensitive information, \textit{this is the best solution we have identified at this stage to idenitify bias on pseudonymised data}, and we encourage any discussion on better approximations. This proxy is reported as binary for simplicity yet we acknowledge gender is not necessarily binary.
Although our research focuses on gender, this should not diminish the importance of investigating other types of algorithmic discrimination.
While this dataset provides important application of fairness-aware algorithms in a high-risk domain, there are several fundamental limitation that can not be addressed easily through data collection or curation processes. These limitations include historical bias that affect a positive outcome for a given user, as well as the impossibility to verify how close the gender-proxy is to the real gender value. Additionally, there might be bias due to the market unfairness that we explained in Section~\ref{subsec:market_unfairness}. Such limitations and possible ethical concerns about the task should be taken into account while drawing conclusions from the research using this dataset. Readers should not interpret summary statistics of this dataset as ground truth but rather as \textit{characteristics of the dataset} only. Additional limitation comes from identifying the \texttt{senior} position label, as the definition of what constitutes a "senior" position can be subjective. We acknowledge that this method may introduce some noise, particularly if job titles are unconventional or if errors occur in categorization. Finally, we remark that in the dataset we assume that each \texttt{user\_id} represents a single user; however, we acknowledge that multiple users could share one device, potentially affecting the user features. Additionally, \texttt{user\_id}'s are based on the company’s identification technology, which means that a single user could have multiple \texttt{user\_id}'s across different browsing sessions. This is one of the complexities inherent in real-world data, particularly in online advertising. Such biases can influence model training, bias evaluation, and bias mitigation efforts.

\section{Empirical observations}

\begin{figure}[ht]
    \centering
    \includegraphics[width=0.32\textwidth]{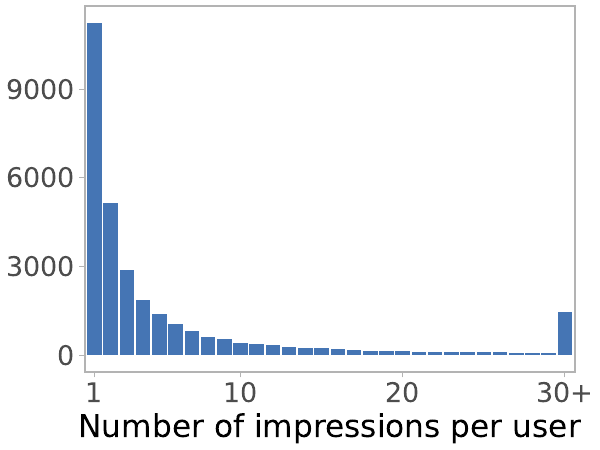}
    \includegraphics[width=0.32\textwidth]{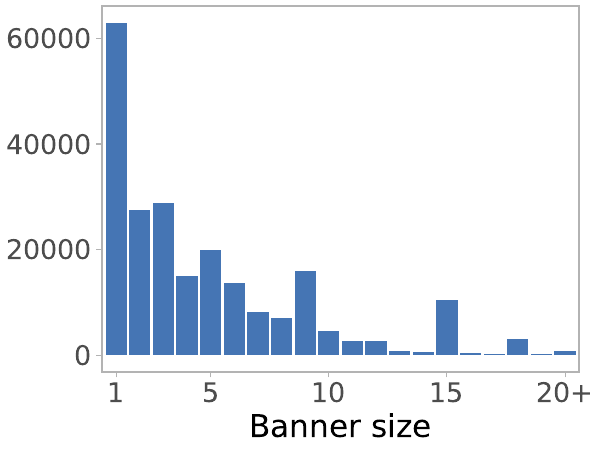}
    \includegraphics[width=0.32\textwidth]{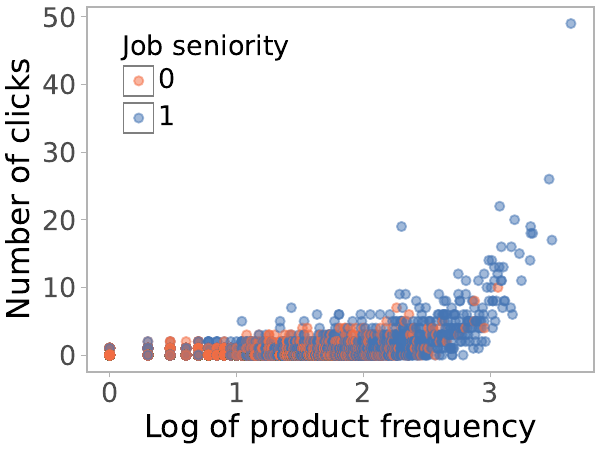}
    \caption{Examples of some feature statistics in FairJob dataset: number of impressions per user and banner size have long tail phenomenon (two plots on the left). The products have popularity bias (right plot), i.e. some products have much higher or lower than average number of clicks with senior job ads having more clicks on average.}
    \label{fig:data_statistics_main}
\end{figure}

\paragraph{Challenges.} The first challenge comes from handling the different types of data that are common in tables, the \textit{mixed-type columns}: there are both numerical and categorical features that have to be embedded~\citep{Gorishniy2021Revisiting,gorishniy2022embeddings,Grinsztajn2022why,shwartz2022tabular, matteucci2023benchmark}. In addition, some of the features have long-tail phenomenon and products have popularity bias, see Figure~\ref{fig:data_statistics_main}. Our datasets contains more than 1,000,000 lines, while current high-performing models are under-explored in \textit{scale}, e.g.  the largest datasets in \citet{Grinsztajn2022why} are only 50,000 lines, while in ~\citet{Gorishniy2021Revisiting,gorishniy2022embeddings} only one dataset surpasses 1,000,000 lines. Additional challenge comes from \textit{strongly imbalanced data}: the positive class proportion in our data is less than 0.007 that leads to challenges in training robust and fair machine learning models~\citep{jesus2022turning,yang2024balanced}. In our dataset there is no significant
imbalances in demographic groups users regarding the protected attribute (both genders are sub-sampled with 0.5 proportion, female profile users were shown less job ad with 0.4 proportion and slightly less senior position jobs with 0.48 proportion), however, there could be a hidden effect of a bias that we discussed in Section~\ref{sec:fairness_in_advertising}. This poses a problem in accurately assessing model performance~\citep{van2024can}. More detailed statistics and exploratory analysis are referred to the supplemental material. 

\paragraph{Baselines.} We choose two baseline regimes: (i) unfair, that uses all attributes for training, including the protected one; and (ii) unaware, that corresponds to fairness through unawareness, i.e. using all attributes during training except the protected one. We train (i) a \texttt{Dummy} classifier in the unaware regime to obtain the first baseline and (ii) \texttt{XGB} in two regimes (unaware and unfair) to achieve a more reasonable baseline performance, see Table~\ref{table:xgb_perf}. We notice that the \texttt{Dummy} classifier is perfectly fair with respect to demographic parity \texttt{DP} which is reasonable since it did not learn the dependence between features at the label at all, as can be seen from the negative log-likelihood \texttt{NLLH} results. However, the utility metrics $U$ and $\tilde U$ of \texttt{Dummy} do not differ much from \texttt{XGB} which is due to very strong imbalance in the data. 
We remark that $\tilde U$ is expectedly higher for fairer (unaware) models, while $U$ is better for the unfair model. There is a slight difference in terms of \texttt{NLLH} and \texttt{AUC-ROC} for \texttt{XGB} models, and higher $U$ corresponds to higher \texttt{AUC-ROC}. We refer to Appendix~\ref{section:experiments_appendix} and~\ref{sec:reproducibility_appendix} for more details. 

\begin{table}[!ht]
\centering
\caption{Performance comparison for single simulation of Dummy classifier (unaware) and XGBoost (unaware and unfair) with 100 trials for tuning.}
\label{table:xgb_perf}
\begin{tabular}{cccccc}
\toprule
 & \texttt{NLLH} $\downarrow$ & \texttt{AUC} $\uparrow$ & \texttt{DP} $\downarrow$ & $U$ $\uparrow$ & $\tilde U$ $\uparrow$ \\
\midrule
\texttt{Dummy} unaware & 0.69239 & 0.50000 & \textbf{0.00000} & 0.01009 & 0.01245 \\
\texttt{XGB} unaware & \textbf{0.05491} & 0.75787 & 0.00278 & 0.01017 &  \textbf{0.01276} \\
\texttt{XGB} unfair & 0.05736 & \textbf{0.76201} & 0.00323 & \textbf{0.01037} &  0.01236  \\
\bottomrule
\end{tabular}
\end{table}

\paragraph{Fair regime.} 
To study a possible trade-off between utility and fairness, we use in-processing fairness methods of adding a fairness-inducing penalty to a loss function $\mathcal{L}( \hat{Y},Y)$ during training~\citep{kamishima2011fairness}:
\begin{equation}
    \hat{Y} = \arg\min \ \ \mathcal{L}( \hat{Y},Y) + \lambda \cdot \text{Penalty}(\hat{Y},Y,A),
\end{equation}
where parameter $\lambda$, or \texttt{fairness\_multiplier}, controls the trade-off between the model's predictive accuracy $\mathcal{L}( \hat{Y},Y)$ and fairness. Adjustments on $\lambda$ allows to control the importance of fairness relative to accuracy.
These methods remove the influence of protected attribute on the model's output without restrictions on  the data~\citep{kamishima2011fairness,bechavod2017penalizing,mary2019fairness}. We implement a penalty based on the approach described in \citet{bechavod2017penalizing}.

\begin{figure}[]
    \centering
    \includegraphics[width=\textwidth]{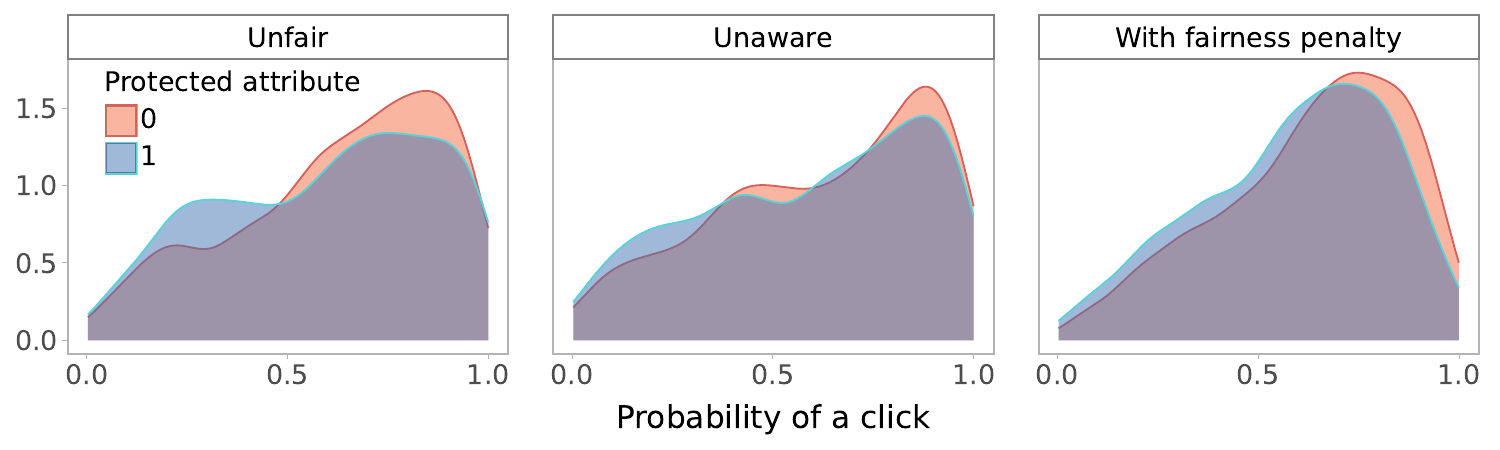}
    \caption{Probability density distributions of click for different values of the protected attribute of three models trained in different ways: (i) \textit{unfair} -- with a protected attribute included as a feature during training, (ii) \textit{unaware} -- corresponds to fairness through unawareness, (iii) trained \textit{with fairness penalty} as a bias mitigation technique. }
    \label{fig:probs}
\end{figure}

We train a logistic regression in the two baseline regimes and compare the results of these algorithms with a (iii) fair model that is trained without protected attribute with an additional fairness-inducing penalty~\citep{kamishima2011fairness,bechavod2017penalizing}.
The three models correspond to the situations described in Figure~\ref{fig:causal_graph}. We refer all the reproducibility details and additional experiments to Appendix~\ref{section:experiments_appendix} and~\ref{sec:reproducibility_appendix}. The resulted prediction can be visually compared in Figure~\ref{fig:probs}.

\begin{figure}[!ht]
    \centering
\includegraphics[width=\textwidth]{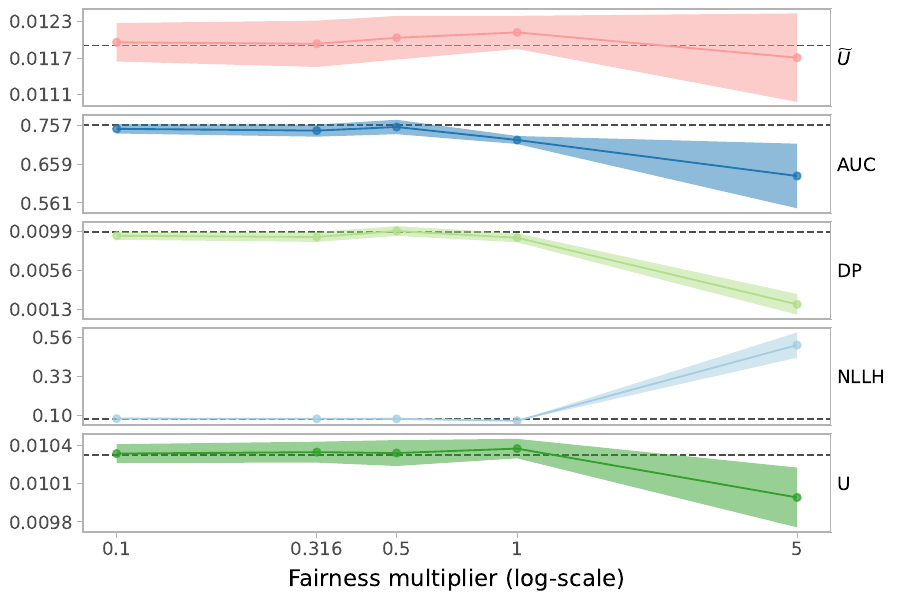}
    \caption{The trade-off between performance and fairness metrics for the logistic regression model when varying the fairness multiplier from 0.1 to 5.0, the variance is reported based on 10 iterations. The dashed line represents unaware logistic regression. }
    \label{fig:lr_fair_metrics}
\end{figure}

\paragraph{Fairness-utility trade-off.} We illustrate the possible trade-off between performance and fairness metrics for the logistic regression model when varying \texttt{fairness\_multiplier}, see Figure~\ref{fig:lr_fair_metrics}. We notice that for positive \texttt{fairness\_multiplier}, \texttt{DP} improves, while \texttt{NLLH} degrades. For \texttt{fairness\_multiplier = 0.5} and \texttt{1.0} we notice slight improvements in utility metrics, especially $\tilde U$, with respect to the unfair model represented as a dashed line.   

Additionally, we propose to restraint an access to the protected attribute and study the trade-off when we train the model on the whole train set but add fairness penalty only for some percentage of train set. In some scenarios, we could see improvements in fairness without sacrificing the overall performance. The loss in accuracy due to the imposed fairness constraints is often small as also noted in other works~\citep{celis2019classification}.
We explore bias correction techniques tailored to address data limitations and preserve utility in large-scale. We demonstrate how prioritizing fairness in AI not only benefits users by fostering inclusivity but also contributes to the long-term success and ethical integrity of companies. 

These findings suggest that the trade-off relationship between accuracy and fairness is context-dependent. It highlights the need for further research to better understand the conditions under which the accuracy fairness trade-off arises and identify strategies to mitigate or overcome it.

\section{Conclusion}

Addressing bias in AI goes beyond mere compliance with legal frameworks like the AI Act; it necessitates proactive measures to detect, prevent, and mitigate biases. 
Drawing from real-world challenges faced by industries, 
we highlight the limitations of existing bias mitigation strategies, particularly in environments where access to sensitive user attributes is restricted. 
We encourage other authors and practitioners to experiment with different AI or Fair AI algorithms on this dataset. We argue that specific problems can often be generalized to broader contexts. For example, if our dataset helps identify a method that effectively balances fairness and utility, this method could potentially be applicable to other recommendation systems across various domains. We expect that with this work, the quality of evaluation of novel AI methods increases, potentiating the development of the area, see more details in the broader impact section in Appendix~\ref{section:broader_impact}. Additionally, we hope it encourages other similar relevant datasets to be published from other authors and institutions.

\section{Acknowledgements}
Federico Pavone has received funding from the European Union's Horizon 2020 research and innovation programme under the Marie Skłodowska-Curie grant agreement No 101034255. We also thank Martin Bompaire, David Rhode and Andre Cunha for helpful discussions.

%%%%%%%%%%%%%%%%%%%%%%%%%%%%%%%%%%%%%%%%%%%%%%%%%%%%%%%%%%%%
\section*{Checklist}

%%% BEGIN INSTRUCTIONS %%%
% The checklist follows the references.  Please
% read the checklist guidelines carefully for information on how to answer these
% questions.  For each question, change the default \answerTODO{} to \answerYes{},
% \answerNo{}, or \answerNA{}.  You are strongly encouraged to include a {\bf
% justification to your answer}, either by referencing the appropriate section of
% your paper or providing a brief inline description.  For example:
% \begin{itemize}
%   \item Did you include the license to the code and datasets? \answerYes{See Section~\ref{gen_inst}.}
%   \item Did you include the license to the code and datasets? \answerNo{The code and the data are proprietary.}
%   \item Did you include the license to the code and datasets? \answerNA{}
% \end{itemize}
% Please do not modify the questions and only use the provided macros for your
% answers.  Note that the Checklist section does not count towards the page
% limit.  In your paper, please delete this instructions block and only keep the
% Checklist section heading above along with the questions/answers below.
%%% END INSTRUCTIONS %%%

\begin{enumerate}

\item For all authors...
\begin{enumerate}
  \item Do the main claims made in the abstract and introduction accurately reflect the paper's contributions and scope?
    \answerYes
  \item Did you describe the limitations of your work?
    \answerYes Section 4, paragraph "Limitations and interpretation". 
  \item Did you discuss any potential negative societal impacts of your work?
    \answerYes 
  \item Have you read the ethics review guidelines and ensured that your paper conforms to them?
    \answerYes 
\end{enumerate}

\item If you are including theoretical results...
\begin{enumerate}
  \item Did you state the full set of assumptions of all theoretical results?
    \answerYes
	\item Did you include complete proofs of all theoretical results?
    \answerYes
\end{enumerate}

\item If you ran experiments (e.g. for benchmarks)...
\begin{enumerate}
  \item Did you include the code, data, and instructions needed to reproduce the main experimental results (either in the supplemental material or as a URL)?
    \answerYes Section Reproducilibity in supplemental meterial
  \item Did you specify all the training details (e.g., data splits, hyperparameters, how they were chosen)?
    \answerYes Section Reproducilibity in supplemental meterial
	\item Did you report error bars (e.g., with respect to the random seed after running experiments multiple times)?
    \answerYes We repeated experiments 10 times with random seeds. 
	\item Did you include the total amount of compute and the type of resources used (e.g., type of GPUs, internal cluster, or cloud provider)?
    \answerYes Section Reproducilibity in supplemental meterial
\end{enumerate}

\item If you are using existing assets (e.g., code, data, models) or curating/releasing new assets...
\begin{enumerate}
  \item If your work uses existing assets, did you cite the creators?
    \answerYes
  \item Did you mention the license of the assets?
    \answerYes
  \item Did you include any new assets either in the supplemental material or as a URL?
    \answerYes
  \item Did you discuss whether and how consent was obtained from people whose data you're using/curating?
    \answerYes
  \item Did you discuss whether the data you are using/curating contains personally identifiable information or offensive content?
    \answerYes
\end{enumerate}

\item If you used crowdsourcing or conducted research with human subjects...
\begin{enumerate}
  \item Did you include the full text of instructions given to participants and screenshots, if applicable?
    \answerNA{}
  \item Did you describe any potential participant risks, with links to Institutional Review Board (IRB) approvals, if applicable?
    \answerNA{}
  \item Did you include the estimated hourly wage paid to participants and the total amount spent on participant compensation?
    \answerNA{}
\end{enumerate}

\end{enumerate}

%%%%%%%%%%%%%%%%%%%%%%%%%%%%%%%%%%%%%%%%%%%%%%%%%%%%%%%%%%%%

\section{Checklist for the dataset track}

\begin{enumerate}

\item  \textcolor{gray}{Submission introducing new datasets must include the following in the supplementary materials:}
\begin{enumerate}
  \item \textcolor{gray}{Dataset documentation and intended uses. Recommended documentation frameworks include datasheets for datasets, dataset nutrition labels, data statements for NLP, and accountability frameworks.} A: Available on the dataset page and in the text in paragraph "License and intended use". 
  \item \textcolor{gray}{URL to website/platform where the dataset/benchmark can be viewed and downloaded by the reviewers.} A: The dataset is hosted on HuggingFace dataset API: FairJob, the other information can be found in supplemental material. 
  \item \textcolor{gray}{URL to Croissant metadata record documenting the dataset/benchmark available for viewing and downloading by the reviewers. You can create your Croissant metadata using e.g. the Python library available here: https://github.com/mlcommons/croissant}
  The dataset viewer automatically generates the metadata in Croissant format (JSON-LD) for every dataset on the Hugging Face Hub. It lists the dataset's name, description, URL, and the distribution of the dataset as Parquet files, including the columns' metadata. The Croissant metadata is available for all the datasets that can be converted to Parquet format.
  \item \textcolor{gray}{Author statement that they bear all responsibility in case of violation of rights, etc., and confirmation of the data license.} A: the statement is in the supplemental material. 
  \item \textcolor{gray}{Hosting, licensing, and maintenance plan. The choice of hosting platform is yours, as long as you ensure access to the data (possibly through a curated interface) and will provide the necessary maintenance.} A: all the details are in supplemental material and on the Hugging Face dataset webpage. 
\end{enumerate}

\item  \textcolor{gray}{To ensure accessibility, the supplementary materials for datasets must include the following:
\begin{enumerate}
  \item Links to access the dataset and its metadata. This can be hidden upon submission if the dataset is not yet publicly available but must be added in the camera-ready version. In select cases, e.g when the data can only be released at a later date, this can be added afterward. Simulation environments should link to (open source) code repositories.
  \item The dataset itself should ideally use an open and widely used data format. Provide a detailed explanation on how the dataset can be read. For simulation environments, use existing frameworks or explain how they can be used.
  \item Long-term preservation: It must be clear that the dataset will be available for a long time, either by uploading to a data repository or by explaining how the authors themselves will ensure this.
  \item Explicit license: Authors must choose a license, ideally a CC license for datasets, or an open source license for code (e.g. RL environments).
  \item Add structured metadata to a dataset's meta-data page using Web standards (like schema.org and DCAT): This allows it to be discovered and organized by anyone. If you use an existing data repository, this is often done automatically.
  \item Highly recommended: a persistent dereferenceable identifier (e.g. a DOI minted by a data repository or a prefix on identifiers.org) for datasets, or a code repository (e.g. GitHub, GitLab,...) for code. If this is not possible or useful, please explain why.
\end{enumerate}}
A: All the dataset details including links, metadata, preservation, license, reproducility details are avaiable in the supplemental material. Licence and intended use information is also stated in the main paper text. 

\item \textcolor{gray}{For benchmarks, the supplementary materials must ensure that all results are easily reproducible. Where possible, use a reproducibility framework such as the ML reproducibility checklist, or otherwise guarantee that all results can be easily reproduced, i.e. all necessary datasets, code, and evaluation procedures must be accessible and documented.} A: All details for the reproducibility are available in the supplemental material. 

\item For papers introducing best practices in creating or curating datasets and benchmarks, the above supplementary materials are not required.
\end{enumerate}

\appendix

\section*{Supplementary materials}

We firstly describe in detail how the dataset FairJob was collected, then provide all the information on the context and features with its statistics. Further, we perform experiments and provide all the steps for the sake of reproducubility. 
The dataset is hosted at \href{https://huggingface.co/datasets/criteo/FairJob}{https://huggingface.co/datasets/criteo/FairJob}. 
Source code for the experiments is hosted at \href{https://github.com/criteo-research/fairjob-dataset/}{https://github.com/criteo-research/FairJob-dataset/}. 

\paragraph{Author statement of responsibility.}
Authors and Criteo bear all responsibility in case of violation of rights and confirmation of the data license.

\section{Dataset information}

\subsection{Data collection and use}

\begin{figure}[!ht]
    \centering
\includegraphics[width=\textwidth]{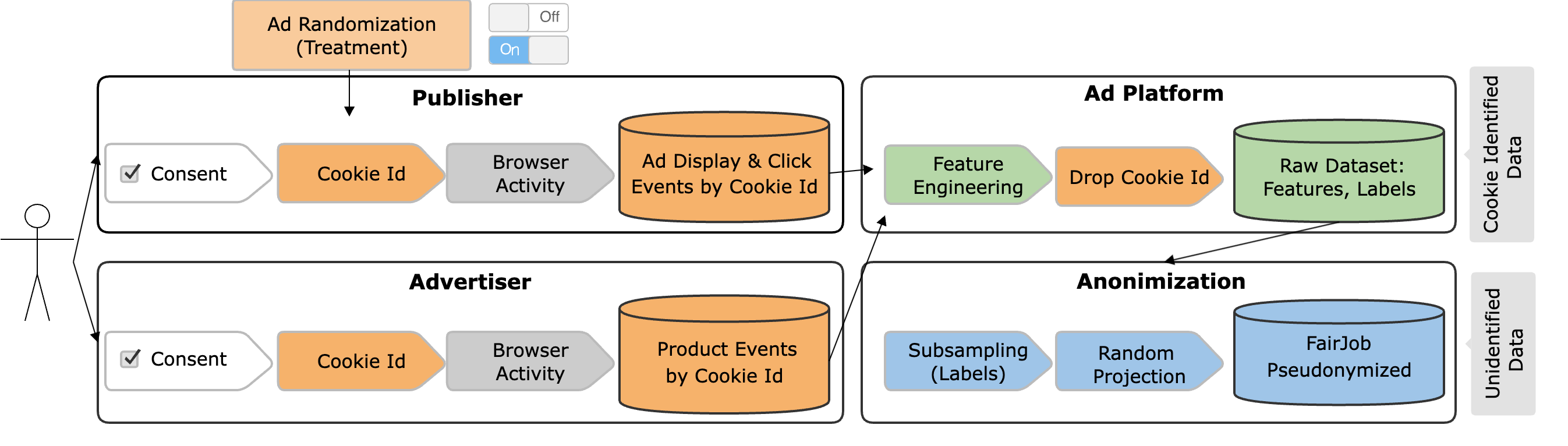}
    \caption{Data collection and processing overview.}
    \label{fig:data_collection}
\end{figure}

As illustrated in Figure~\ref{fig:data_collection}, the process starts with users navigating Publisher and Advertiser websites (typically newspapers and retailer shops respectively). Upon user consent\footnote{In accordance with GDPR and Article 82 of the Data Protection Act that require the provider to ask consent of data subjects if it was reading/writing information to the user's device.}, user information about the events such as visits or product views\footnote{The original data does not contain any "special categories" of personal data listed under Article 9 of the GDPR, processing of which is prohibited, except in limited circumstances set out in Article 9 of the GDPR.} are collected and identified by means of browser cookies. Users are subject to personalized advertising (if the job campaign was chosen and the display opportunity was won) until the end of the data collection period. Subsequently, only won displays coming from Publisher and Advertiser partners are joined by cookie identifier on the AdTech platform to form the raw dataset, dropping cookie ids when they are not needed anymore. 

Further, the data is collected by trackers upon a page call and is commonly stored as a tuple of page and client information, which we call a click.
Several steps are in place to ensure that clicks on ads must result from a human user with genuine interest. This steps are stated in the guidelines that are accepted by publishers\footnote{The guidelines are available on the official webpage: \href{https://www.criteo.com/supply-partner-guidelines/}{https://www.criteo.com/supply-partner-guidelines/}.}. Any method or mechanism that artificially generates clicks or impressions is strictly prohibited, including but not limited to the following:
\begin{itemize}
\item Clicks generated by publishers clicking on their own ads.
\item Repeated clicks on the same ad unit.
\item Publishers encouraging users to click on their ads (examples may include: any language encouraging users to click on ads; ad implementations that may cause a high volume of accidental clicks; monetary or non-monetary incentives or rewards for clicks, etc.).
\item Use of automated clicking tools or traffic sources, robots, or other deceptive software, click spam or click injections.
\item Use of any other artificial mechanism to generate or inflate clicks.
\item Clicks generated outside of the ad surface will not be counted as intended clicks.
\end{itemize}
Additional safeguards are related to clearly stating that the user sees an ad. Any native ad formats displayed on the publisher’s site must be clearly marked as an ad. Ads should not be placed very close to or underneath buttons or any other object such that the placement of the ad interferes with a user’s typical interaction with the Site content or functionalities.

Publishers accept the guidelines which state the following:
\begin{itemize}
\item Publishers may offer their users the opportunity to view ads in exchange for user rewards or incentives, but only if the user is not forced or incentivized to interact with the ad, such as incentivized click.
\item Publishers must not, directly or indirectly, provide incentives to users in exchange for clicks on ads, or make use of any mechanism or monetizable reward to incentivize clicks.
\end{itemize}

To ensure confidentiality, the collected data has been \textit{sub-sampled non-uniformly} to avoid disclosing business metrics. Feature names have been \textit{anonymized}, and their values \textit{randomly projected} to preserve predictive power while rendering the recovery of the original features or user context practically impossible\footnote{The orginal data is still used in the AdTech company, however, re-indentification of the pseudomymized data is impossible due to additional randomization techniques during data anonymization to comply with Article 4(5) of GDPR.}. The dataset does not contain the relevant attributes such as gender; however, it includes a gender proxy which we discuss in detail in the main text.  

\paragraph{License and intended use.} The data is released under the CC-BY-NC-SA 4.0 license which gives liberty to Share and Adapt this data provided that the respect of the Attribution, NonCommercial and ShareAlike conditions. We focus in this paper on algorithmic fairness analysis as a specific use-case in job-seeking contexts, but it does not have to restricted to this. Fairness in job advertising is a crucial step towards ensuring fairness in any advertising campaign involving sensitive topics. The FairJob datasets presents a challenge of training fair and robust models on strongly imbalanced data~\citep{jesus2022turning,yang2024balanced}. While many of the features in the FairJob dataset (such as user features and categorical product features) are not specific to job advertising and can be applied to other domains for improving click-prediction methods, however, the consumer behavior may vary significantly from a job seeker's behavior. 
Additionally, the dataset can be used as a baseline for improving deep learning models on tabular data and to study methods on embedding creation for numerical and categorical features as done by~\citet{Gorishniy2021Revisiting,gorishniy2022embeddings,Grinsztajn2022why,shwartz2022tabular, matteucci2023benchmark}. Another possible usage is to explore privacy-preserving techniques for tabular data as in~\citet{donhauser2024privacy}. Moreover, this dataset can be used to improve the generation techniques on tabular data, as described in~\citet{jesus2022turning,van2024tabular}.
We compare FairJob dataset details to other commonly used open-source tabular datasets further in Section~\ref{sec:tabular_datasets}.

\subsection{User privacy protection}

The original dataset does not include any data that directly identifies a user, such as names, postal addresses, or email addresses in plain text, so the \textit{original data is pseudonymized}. However, due to possible uniqueness in high-dimensional datasets, it has been shown that some records in pseudonymized datasets might be re-identifiable by an adversary. We describe a possible re-identification process by an adversary and demonstrate that it is practically impossible.

In the FairJob dataset, there are both user and product features. We assume that only user features could potentially be used to identify a user. The product features are categorical, based on internal company categories across different products and brands, or on information provided by clients, and thus are not related to users.

We can suppose that an adversary runs a company that collects user browsing features and, therefore, has an auxiliary dataset with user information. The adversary's goal is to identify a user from the FairJob dataset. Since all features are anonymized, the adversary cannot directly match the features and must first align FairJob feature distributions with the adversary’s company's feature distributions and then find a unique association from a FairJob record to the adversary’s company record, thereby identifying the user.
However, since we anonymized, added noise, and standardized continuous user features, re-identification becomes impossible as the original scale is not provided. The only features of potential interest to the adversary are two categorical user features.

The categorical user features are based on internal company categories, which are unlikely to be known to the adversary due to business confidentiality, making it impossible for the adversary to reconstruct them. Additionally, features \texttt{cat0} and \texttt{cat1} have the same cardinality, and their distributions are not significantly different, making them indistinguishable and effectively reducing the number of possible unique variations. Most importantly, both features \texttt{cat0} and \texttt{cat1} have a cardinality of 9, meaning the tuples (\texttt{cat0}, \texttt{cat1}) can have a maximum of 81 variations among 1 million records. In FairJob, each present variation of (\texttt{cat0}, \texttt{cat1}) corresponds to at least 10 records, making it impossible for an instance (\texttt{cat0}, \texttt{cat1}) to serve as a pseudonym for a user, i.e., user re-identification based on these two features is not possible.

All these measures make it \textit{impossible to re-identify users} from the FairJob dataset.

\subsection{Dataset detailed description}
The dataset contains pseudononymized users' context and publisher features that was collected from a job targeting campaign ran for 5 months by Criteo AdTech company.  Each line represents a product that was shown to a user. Each user has an impression session where they can see several products at the same time. Each product can be clicked or not clicked by the user. The dataset consists of  1072226 rows and 55 columns:
\begin{itemize}
   \item \texttt{user\_id} is a unique identifier assigned to each user. This identifier has been anonymized and does not contain any information related to the real users. 
    \item \texttt{product\_id} is a unique identifier assigned to each product, i.e. job offer. 
    \item \texttt{impression\_id} is a unique identifier assigned to each impression, i.e. online session that can have several products at the same time. An impression, or successful online session, is recorded when a product banner (or multiple banners) is loaded and begins to render on the publisher's page. An impression can subsequently lead to a click event.
   \item \texttt{cat0}, $\dots$, \texttt{cat1} are anonymized categorical user features. 
    \item \texttt{cat2}, $\dots$, \texttt{cat12} are anonymized categorical product features. 
    \item \texttt{num13}, $\dots$, \texttt{num47} are anonymized numerical user features. 
    % (in the paper, we mistakenly described 6 continuous user features; these should be 2 user and 4 product features, which has been corrected)
    \item \texttt{protected\_attribute} is a binary feature that describes user gender proxy, i.e. female is~0, male is~1. The detailed description on the meaning can be found in the main paper. 
    \item \texttt{senior} is a binary feature that describes the seniority of the job position, i.e. an assistant role is 0, a managerial role is 1. It includes both managerial and individual contributor (IC) roles, as our aim was to encompass any senior-level positions. This feature was created during the data processing step from the product title feature: if the product title contains words describing managerial role (e.g. 'president', 'ceo', and others), it is assigned to 1, otherwise to 0.
    \item \texttt{rank} is a numerical feature that corresponds to the positional rank of the product on the display for given \texttt{impression\_id}. Usually, the position on the display creates the bias with respect to the click: lower rank means higher position of the product on the display. 
    \item \texttt{displayrandom} is a binary feature that equals 1 if the display position on the banner of the products associated with the same \texttt{impression\_id} was randomized. The click-rank metric should be computed on \texttt{displayrandom = 1} to avoid positional bias. 
    \item \texttt{click} is a binary feature that equals 1 if the product \texttt{product\_id} in the impression \texttt{impression\_id} was clicked by the user \texttt{user\_id}. 
\end{itemize}
Figure~\ref{fig:data_statistics} illustrates distributions of some features: number of impressions per user, number of products per user and banner size have long tail phenomenon (plots of the upper row). The products have popularity bias (lower plot), i.e. some products have much higher or lower than average number of clicks with senior job ads having more clicks on average, and position bias, i.e. increased number of clicks per lower rank and almost no clicks in the highest ranks (right plot on the lower row).

\begin{figure}[!ht]
    \centering
\includegraphics[width=\textwidth]{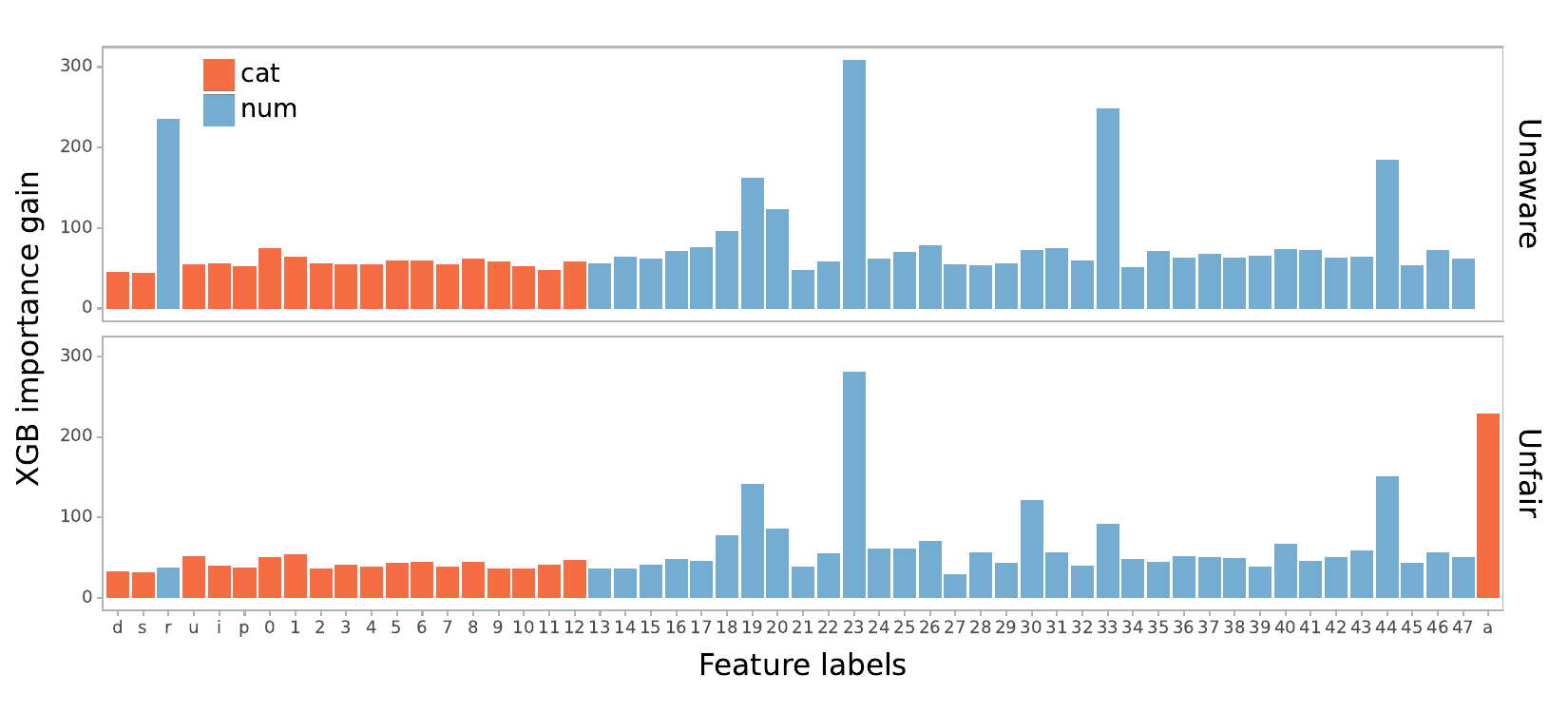}
    \caption{XGBoost importance gain per feature when trained in \textit{unaware} way (without protected attribute as a feature -- upper plot) and \textit{unfair} way (with protected attribute as a feature -- lower plot). Label \texttt{d} corresponds to \texttt{displayrandom}, \texttt{s} to \texttt{senior}, \texttt{r} to \texttt{rank}, \texttt{u} to \texttt{user\_id}, \texttt{i} to \texttt{impression\_id}, \texttt{p} to \texttt{product\_id}, labels 0 to 47 to categorical (\texttt{cat0}, \dots, \texttt{cat12}) and numerical (\texttt{num13}, \dots, \texttt{num47}) features, respectively, and only on the lower plot \texttt{a} to \texttt{protected\_attribute}. }
    \label{fig:xgb_imp}
\end{figure}

\begin{figure}[ht]
    \centering
    \includegraphics[width=0.32\textwidth]{impressions_per_user.pdf}
     \includegraphics[width=0.32\textwidth]{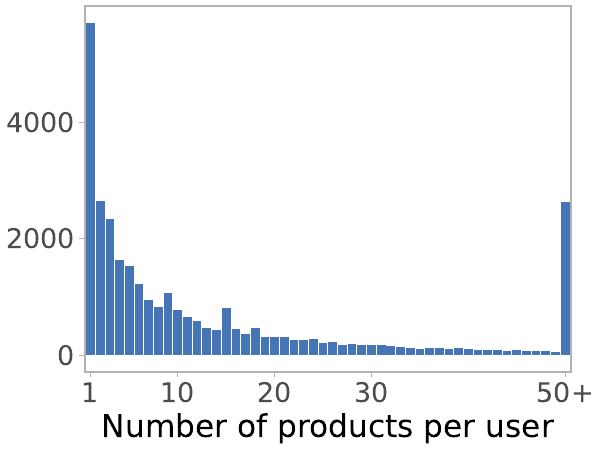}
    \includegraphics[width=0.32\textwidth]{banner_size.pdf}
    \includegraphics[width=0.32\textwidth]{prodfreq_vs_clicks.pdf}
    \includegraphics[width=0.32\textwidth]{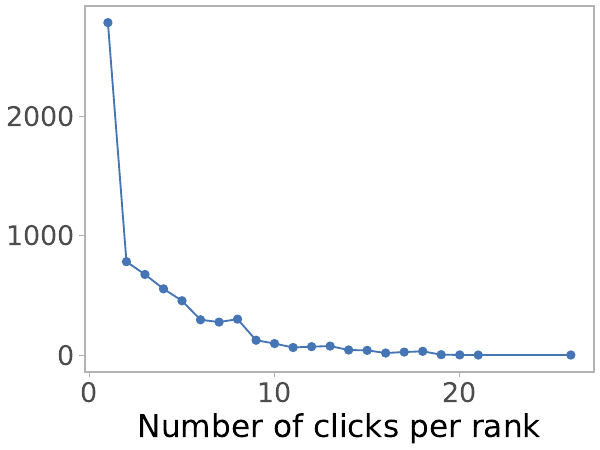}
    \caption{Examples of some feature statistics in FairJob dataset: number of impressions per user, number of products per user and banner size have long tail phenomenon (plots of the upper row). The products have popularity bias (left plot on the lower row), i.e. some products have much higher or lower than average number of clicks with senior job ads having more clicks on average. We observe a position bias due to increased number of clicks per rank and almost no clicks in the highest ranks (right plot on the lower row).}
    \label{fig:data_statistics}
\end{figure}

Figure~\ref{fig:xgb_imp} represents the feature importance according to an importance gain of XGBoost trained in two different ways: (i) \textit{unaware} without protected attribute as a feature and (ii) \textit{unfair} way with protected attribute as a feature. Their performance is reported in Table~\ref{table:xgb_perf}. We notice that the feature \texttt{rank} has a high importance, however, it gets replaced by the \texttt{protected\_attribute} feature in the unfair regime. There most impactful features for both models are \texttt{num23}, \texttt{num33}, \texttt{num43}, \texttt{num18} and \texttt{num19}, apart from \texttt{rank} for the unaware model and \texttt{protected\_attribute} for the unfair one.   

\begin{figure}[!ht]
\centering
    \includegraphics[width=0.9\textwidth]{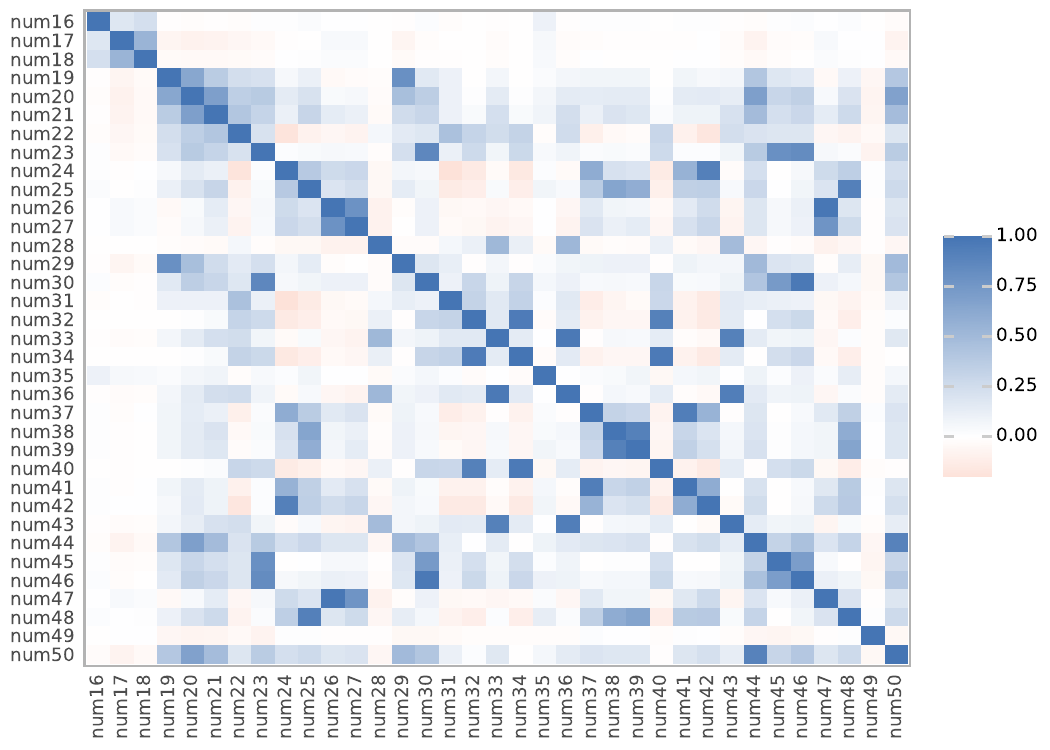}
    \caption{Correlations between numerical features in the \textbf{FairJob} dataset.}
    \label{fig:num_correlations}
\end{figure}

\begin{figure}[!ht]
\centering
    \includegraphics[width=0.9\textwidth]{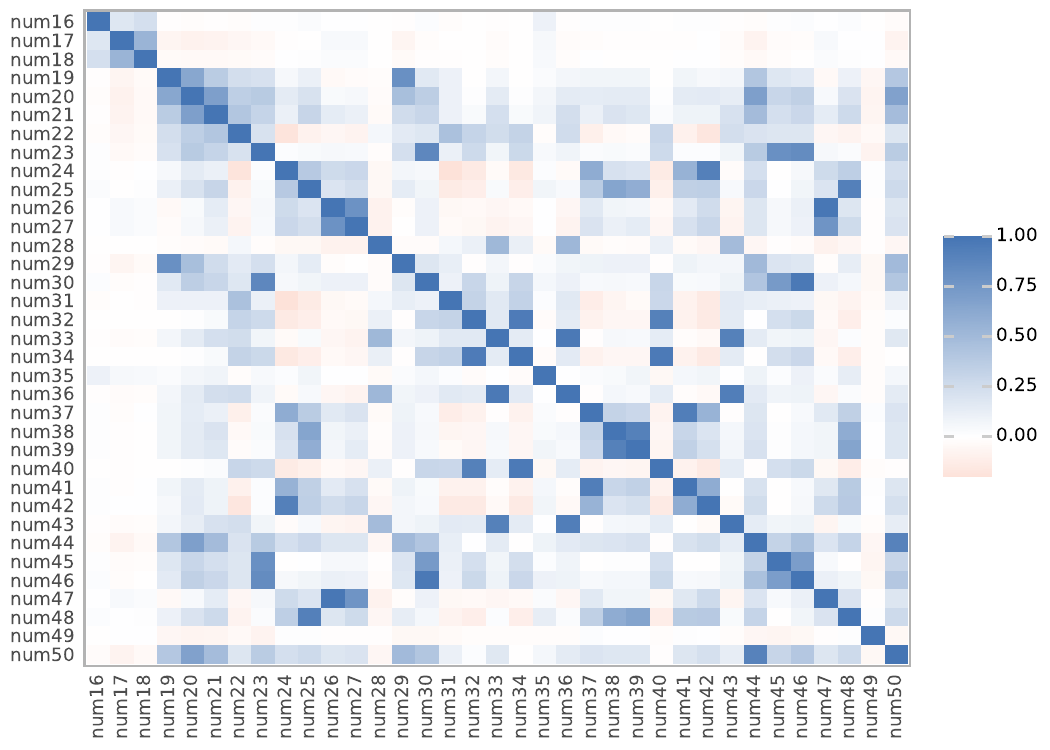}
    \caption{Correlations between numerical features in the \textbf{original} dataset.}
    \label{fig:original_num_correlations}
\end{figure}

\begin{figure}[!ht]
\centering
    \includegraphics[width=0.9\textwidth]{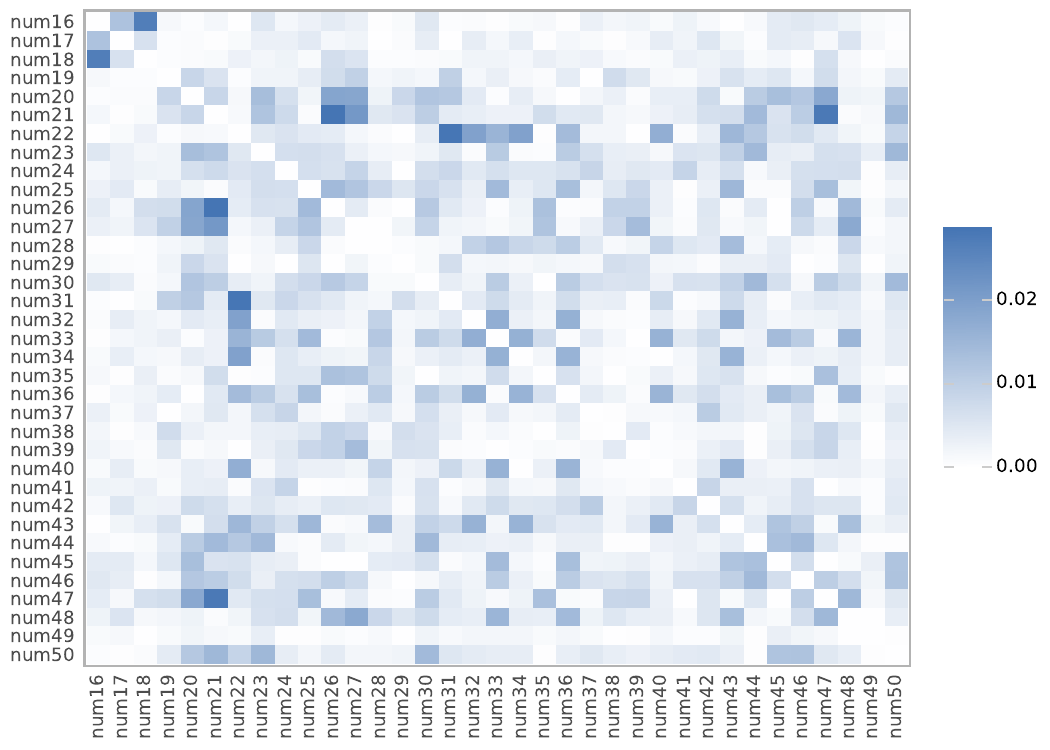}
    \caption{\textbf{Difference} in correlations between numerical features in the \textbf{original} and \textbf{FairJob} dataset.}
    \label{fig:diff_num_correlations}
\end{figure}

In Figures~\ref{fig:num_correlations} and \ref{fig:original_num_correlations} we plot correlation matrix between numerical features in the original dataset and open-sourced dataset FairJob, where we added noise for anonymization. We observe that correlations are generally slightly weaker for most features, when observing the anonymized data, that can be better seen in Figure~\ref{fig:diff_num_correlations}, where we plot difference between the correlations. This fact comes naturally from the added noise, however, lower correlation values might translate into higher classification difficulty.

\begin{table}[!ht]
\caption{Categorical features cardinalities.}
\label{table:cat_features}
\centering
{\footnotesize
\begin{tabular}{cc@{\ \ }c@{\ \ }c@{\ \ }c@{\ \ }c@{\ \ }c@{\ \ }c@{\ \ }c@{\ \ }c@{\ \ }c@{\ \ }c@{\ \ }c@{\ \ }c@{\ \ }c@{\ \ }c@{\ \ }c
}
\toprule
feature & \texttt{cat0} & \texttt{cat1} & \texttt{cat2} & \texttt{cat3} & \texttt{cat4} & \texttt{cat5} & \texttt{cat6} & \texttt{cat7} & \texttt{cat8} & \texttt{cat9} & \texttt{cat10} & \texttt{cat11} & \texttt{cat12} \\ \midrule
cardinality & 9 & 9 & 1025 & 98 & 122 & 1296 & 2492 & 3183 & 3541 & 2879 & 2314 & 1436 & 912\\ \bottomrule
\end{tabular}
}
\end{table}
\begin{table}[!ht]
\caption{Statistics for index features.}
\label{table:index_features}
\centering
{\footnotesize
\begin{tabular}{cc@{\ \ }c@{\ \ }c}
\toprule
feature & \texttt{user\_id} & \texttt{impression\_id} & \texttt{product\_id} \\ \midrule
cardinality & 30361 & 224898 & 57355 \\ \bottomrule
\end{tabular}
}
\end{table}

\begin{table}[!ht]
\caption{Statistics for binary features (out of 1072226 rows).}
\label{table:binary_features}
\centering
{\footnotesize
\begin{tabular}{cc@{\ \ }c@{\ \ }c@{\ \ }c}
\toprule
feature & \texttt{protected\_attribute} & \texttt{senior} & \texttt{displayrandom} & \texttt{click}\\ \midrule
positive & 536113 (50\%) & 713659 (66.6\%) & 105869 (9.9\%) & 7489 (0.7\%)\\ \bottomrule
\end{tabular}
}
\end{table}


\subsection{Dataset statistics}
Table~\ref{table:cat_features}, \ref{table:index_features} and \ref{table:binary_features} demonstrate the available statistics for categorical features cardinalities, index features and binary features respectively. Additionally, we provide statistics for selection bias of the job campaign that comes from the Criteo AdTech company (outside of FairJob dataset) in Table~\ref{table:selection_bias}, which we take into account further to compute utility metrics.

\begin{table}[!ht]
\caption{Statistics for job campaign selection bias: we see that while female (internet)-population is slightly larger than male, the female population in the job campaign is much smaller, leading to selection bias of the job campaign -- male users are almost twice more picked than female users.}
\label{table:selection_bias}
\centering
{\footnotesize
\begin{tabular}{cc@{\ \ }c@{\ \ }c}
\toprule
population & advertising & job campaign & ratio \\ \midrule
female & 53.6\% & 39.2\% & 0.73 \\ 
male & 46.4\% & 60.7\% &  1.31 \\ \bottomrule
\end{tabular}
}
\end{table}

We provide the detailed statistics on clicks, job seniority and the protected attribute in the dataset in Table~\ref{table:click_senior_attribute}. From this table we can compute probabilities of events. For example, Table~\ref{table:protected_attribute_given_seniority} shows that if a job ad is about a senior position, there is slightly higher chance that it will be shown to a male user than female (by 3.36\%), while an assistant job ad is more likely to be shown to a female user than male (by 6.68\%). 
In contrast, as can be seen in Table~\ref{table:click_senior_given_attribute}, there is almost no difference in clicks for senior ads given the protected attribute (both 0.49\%) and a job add is slightly more likely to be clicked if shown to a female user (by 0.06\%). 

\begin{table}[!ht]
    \centering
\caption{Observed frequencies for clicks, senior job ads, and protected attribute}
\label{table:click_senior_attribute}
\begin{tabular}{lccccc}
\toprule
 & \multicolumn{2}{c}{not clicked} & \multicolumn{2}{c}{clicked}  & \\
\midrule
 & non-senior & senior & non-senior & senior & all \\
\midrule
female & 189982 & 342221 & 1274 & 2636 & 536113 \\
male & 166394 & 366140 & 917 & 2662 & 536113 \\ \midrule
all & 356376 & 708361 & 2191 & 5298 & 1072226 \\
\bottomrule
\end{tabular}
\end{table}

\begin{table}[!ht]
\centering
\caption{Observed frenquencies for user's protected attribute given the seniority of the shown job ad.}
\label{table:protected_attribute_given_seniority}
\begin{tabular}{ccc}
\toprule
& non-senior & senior  \\
\midrule
female  & 0.533390 & 0.483224  \\
male  & 0.466610 & 0.516776 \\
% \midrule
% all & 1.0 & 1.0 \\ 
\bottomrule
\end{tabular}
\end{table}

\begin{table}[!ht]
\centering
\caption{Observed joint frenquencies for clicks, senior job ads, conditional on the protected attribute.}
\label{table:click_senior_given_attribute}
\begin{tabular}{ccccc}
\toprule
 & \multicolumn{2}{c}{not clicked} & \multicolumn{2}{c}{clicked} \\
 & non-senior & senior & non-senior & senior \\
\midrule
female & 0.354369 & 0.638337 & 0.002376 & 0.004917 \\
male & 0.310371 & 0.682953 & 0.001710 & 0.004965 \\
\bottomrule
\end{tabular}
\end{table}

\section{Experiments}
\label{section:experiments_appendix}

\subsection{Metrics}
\label{sec:metrics}

\paragraph{Fairness metrics. }
To measure fairness of the model, we consider \textit{demographic parity} on senior job opportunities that computes the average difference of predictions given the protected attribute and the job seniority:
\begin{equation}
    \text{DP} (\hat{Y} | A) =  \mathbb{E} \left( \hat{Y} | A = 1, S = 1 \right) -  \mathbb{E} \left( \hat{Y} | A = 0, S = 1 \right). 
\end{equation}
Demographic parity requires equal proportion of positive predictions in each group ("No Disparate Impact") which in our case can be translated as same proportion of shown senior job ads in each group of the protected attribute. Demographic parity can be thought of as a stronger version of the US Equal Employment Opportunity Commission’s ``four-fifths rule'', which requires that the ``selection rate for any race, sex, or
ethnic group [must be at least] four-fifths (4/5) (or eighty percent) of the rate for the group with the highest rate''\footnote{See the Uniform Guidelines on Employment Selection Procedures, 29 C.F.R. §1607.4(D) (2015).}.

\textit{Equalized odds}~\citep{zafar2017fairness} ensures that a machine learning model works equally well for different groups. In case of job ads, equalized odds would force the prediction of both positive and negative outcomes to be the same which might generate more false positive predictions for one group versus others, resulting in worse outcome. 
\textit{Equal of opportunity}~\citep{hardt2016equality} ensures that a machine learning model works equally well only on positive outcomes for different groups which results in capturing the costs of misclassification disparities.

\paragraph{Performance metrics.} % FPR stands for False Positive Rate, FNR for False Negative Rate, PR for the rate of positive classification, 
The loss function is log-loss and report it as \texttt{NLLH} (negative log-likelihood). We also report 
\texttt{AUC} (Area under the ROC Curve) as a description of prediction power on strongly imbalanced data in binary classification problems. Additionally, we consider click rank utility $U$ and product-rank utility for biased data $\tilde U$, proposed in the main text Section~\ref{subsec:recommendation_bias}. 

\subsection{Models.}

\paragraph{Training regimes.} We train models in the following ways: 
\begin{itemize}
    \item \textbf{unfair} --  the model uses protected attribute as a feature in the data, 
    \item \textbf{unaware} -- the model does not use the protected attribute in the data, corresponds to fairness through unawareness, 
    \item \textbf{fair} --  the model does not use the protected attribute in the data and trained with an additional fairness-enforcing penalty in the loss. 
\end{itemize}

\paragraph{Algorithms.} As baseline results, we perform three methods in different regimes: 
\begin{itemize}
    \item \texttt{Dummy} -- classifier based on a single threshold for positive class probability (unaware), 
    \item \texttt{XGB} -- XGBoost algorithm (unfair and unaware), 
    \item \texttt{LR} -- Logistic Regression (unfair, unaware, and fair).
\end{itemize}

\subsection{Possible improvements.} 

\paragraph{Base methods.} We considered XGBoost and Logistic Regression models, but in some cases deep learning models such as Multi-Layer Perceptron, ResNets or Transformers might perform better, especially when focusing on improving mixed embeddings of categorical and numerical features~\citep{gorishniy2022embeddings}. We also stress the challenge of strongly imbalanced classification, which can be further investigated. 

\paragraph{Fairness methods.} Another way to enforce fairness during training is to use an adversarial algorithm, e.g.~\citet{lahoti2020fairness}. Alternatively, the pre-processing corrections are related to the direct modifications of the training set before training. The examples of the methods include separation of observatives $X$ into two subsets $X_{\text{desc}(A)}$ and $X_{\text{non-desc}(A)}$~\citep{zemel2013learning}; transformation of $X$ to some $\tilde X$ so that its factual and counterfactual distributions becomes the same, i.e. $P_{\tilde X} | do(A = a) = P_{\tilde X} | do(A = a')$ for all $a, a'$;  learning "disentangled" representations~\citep{locatello2019fairness}.   
The pre-processing methods might be hard to apply to large and complex data as FairJob and might lead to losing a lot of data and, therefore, performance. However, the pre-processing and in-processing techniques might be combined together to achieve better results. We refer to \citet{hort2023bias} for a recent survey on fairness inducing methods. 

\section{Reproducibility}
\label{sec:reproducibility_appendix}

Source code for the experiments is hosted at \href{https://github.com/criteo-research/fairjob-dataset/}{https://github.com/criteo-research/fairjob-dataset/}.

\paragraph{Resources.}
Experiments were conducted on Criteo internal cluster on instances with a RAM of 500Go and 46 CPUs available and 6 GPUs V100.

\paragraph{Tuning.} We tune each model’s hyperparameters following the procedure and the hyperparameters search spaces from~\citet{gorishniy2022embeddings}. Namely, the best hyperparameters are the ones that perform best on the validation set, so the test set is never used for tuning. For most
algorithms, we use the Optuna library~\citep{akiba2019optuna} to run Bayesian optimization (the Tree-Structured Parzen
Estimator algorithm)~\citep{turner2021bayesian}. We set the number of tuning trails to 50, we use pruning and upper bound the number of data examples for hyperparameter search to 50000 as a compromise between optimality and training time constraint, however, we train the models with the optimal parameters on the whole training set. The amount of data and number of trails for tuning can be increased for the best performance. 

\subsection{Logistic regression details} 

\paragraph{Feature embeddings.} We use a PyTorch~\citep{paszke2019pytorch} module \texttt{Embedding} to compute embeddings for categorical features as a part of model training. Then, we created a mixed embedding layer where we concatenate categorical features embeddings with numerical features. The resulted mixed embedding layer is used as a model input. The embeddings for both categorical and numerical features can be improved, for example, by following the benchmark paper of \citet{gorishniy2022embeddings}. 

\paragraph{Fixed hyperparameters.}
We tested different batch sizes from [1024, 4000, 10000] and fixed batch=1024 as the best performing. Due to strong class imbalance, we oversample positive class examples for batch generation with sampling weights inversely proportional to the observed class frequencies in the data, using PyTorch~\citep{paszke2019pytorch} utility \texttt{WeightedRandomSampler}. We fix number of epochs to \texttt{n\_epochs} = 50.

\paragraph{Tuned hyperparameters.} We tune the following hyperparameters:
\begin{itemize}
    \item \texttt{embedding\_size = UniformInt[4, 8]}, 
    \item \texttt{weight\_decay = LogUniform[1e-6, 1e-4]},
    \item \texttt{scheduler\_step\_size = UniformInt[20, n\_epochs]}, 
    \item \texttt{scheduler\_gamma = LogUniform[1e-2,1]}
\end{itemize}

\paragraph{Fairness parameters.}
We report the results for 
\begin{itemize}
\item \texttt{fairness\_multiplier} $\lambda \in [0.0, 0.1, 0.316, 1.0, 3.0, 5.0]$.
\end{itemize}
For each configuration we performed hyperparameter tuning described above. 

\paragraph{Evaluation.} For each tuned configuration, we run 10 experiments with different random seeds and report the average performance on the test set.

\paragraph{Experiment for density plots.}
The density plots illustrated in the main paper (Figure 5) correspond to the positive predictions densities for logistic regression model trained in \textit{unfair}, \textit{unaware} and \textit{fair} ways. For easy and fast reproducibility, we upper-bound the number of training examples for  hyperparameters tuning to 100000,  \texttt{batch\_size = 10000} and \texttt{number\_of\_epochs = 15}. 
The results are outputs of one simulation of logistic regression models with the following parameters and tuned on the following hyperspaces: 
\begin{itemize}
    \item \texttt{embedding\_size = UniformInt[4, 5]}, 
    \item \texttt{learning\_rate = LogUniform[1e-4, 1e-2]},
    \item \texttt{weight\_decay = LogUniform[1e-6, 1e-4]}
\end{itemize}
For the model with fairness penalty, we use penalty coefficient equal to 3.0.
We want to stress that the aim of this simulation was to provide a descriptive example rather than the optimal model for click classification.
In the repository, we report the best parameters and predictions, as well as the code for the plot generation. 

\subsection{XGBoost details}
\paragraph{Feature embeddings.} We use the Python package for the XGBoost library \citep{chen2016xgboost}, through the Scikit-Learn API \citep{scikit-learn}. We encode the categorical features using the \texttt{TargetEncoder} in Scikit-Learn, learning the encoding on the training set~\citep{micci2001preprocessing}.

\paragraph{Fixed hyperparameters.} 
We use the histogram (\texttt{hist}) tree contruction algorithm in order to speed up the model fit. In order to deal with the imbalance of the classes, we set the \texttt{scale\_pos\_weight} parameter to the ratio between negative and positive class occurrences in the training data, as suggested in the library documentation. We fix number of trials for tuning to \texttt{n\_trials} = 100.

\paragraph{Tuned hyperparameters.}
We tune the following hyperparameters:
\begin{itemize}
    \item \texttt{max\_depth = UniformInt[3,10]},
    \item \texttt{min\_child\_weight = LogUniform[0.0001,100]},
    \item \texttt{subsample = Uniform[0.5,1]},
    \item \texttt{learning\_rate = LogUniform[0.001,1]},
    \item \texttt{colsample\_bytree = Uniform[0.5,1]},
    \item \texttt{reg\_lambda = LogUniform[0.1,10]},
    \item \texttt{gamma = LogUniform[0.001,100]}.
\end{itemize}

\section{Comparison to other tabular datasets}
\label{sec:tabular_datasets}

\subsection{Click-prediction datasets}

We argue that the FairJob dataset is representative enough to evaluate click prediction algorithms, based on three main points:
\begin{itemize}
\item \textbf{real-world data} -- the dataset is based on real-world data and contains all the necessary information. We have provided detailed documentation to support this.
\item \textbf{comparative size} -- the dataset is comparable in size to other widely-used specialized recommendation datasets.
\item \textbf{baseline experiments} -- we performed baseline experiments that produced reasonable results with respect to both performance and the utility-fairness trade-off.
\end{itemize}
The dataset contains pseudonymized users’ context and publisher features collected from a job-targeting campaign run over five months. Therefore, the dataset contains \textit{slices of real-world data}.

To properly compare the FairJob dataset to other click prediction datasets, we searched for the most used datasets for click prediction. We identified the following widely-used datasets for click prediction based on benchmark papers~\citep{zhu2021open,mao2023finalmlp}:
\begin{itemize}
    \item Criteo 2014 Dataset. It consists of ad click data over a week and comprises 26 categorical feature fields and 13 numerical feature fields.
    \item Avazu 2015 Dataset. It contains 10 days of click logs and has a total of 23 fields with categorical features including app id, app category, device id, etc.
\end{itemize}
Both datasets are sampled from real click logs in production and contain tens of millions of samples. However, both lack explicit user and item field information.
In addition to click prediction datasets, other recommendation system problems of similar complexity, such as MovieLens for predicting engagements and Frappe related to app usage, have datasets comparable in size to FairJob~\citep{zhu2021open,mao2023finalmlp}. See the detailed comparison in Table~\ref{tab:click_datasets}.

\begin{table}[!ht]
\caption{Comparison of FairJob to other most frequently used tabular datasets for click prediction and recommendation system problems of similar complexity.}
\label{tab:click_datasets}
\centering
\begin{tabular}{{@{\ } c @{\ }c@{\ }c@{\ }c @{\ }c @{\ }c @{}}}\toprule
\text{dataset} & \text{\ \# rows} & \text{\ \# features} \\ \hline
\text{Criteo} & 45,840,617 & 39 \\ 
\text{Avazu} & 40,428,967 & 22 \\ 
\text{MovieLens} & 2,006,859 & 3 \\ 
\rowcolor{black!20}[0pt][0pt] \text{FairJob} & 1,072,226 & 60 \\ 
\text{Frappe} & 288,609 & 10  \\
\bottomrule
\end{tabular}
\end{table}

We acknowledge that the FairJob dataset does not contain all features typically used for training click prediction models in the industry, due to business confidentiality. However, despite hand-picking the most relevant features, our dataset still contains 60 features, which is the largest number of real features in a dataset for click prediction.

The FairJob dataset does focus on a subproblem of click prediction within the context of job advertising campaigns, where users are pre-selected as having potential job-seeking profiles. Consequently, the dataset size is represented by a smaller number of records specific to this campaign. The click rate of 0.007\% is consistent with the click rates observed in larger-scale datasets.

Furthermore, 1 million records for an unbalanced classification problem seems a reasonable size compared to other tabular datasets, as also stated in the next subsection. Our experiments in Appendix~\ref{section:experiments_appendix} demonstrate that baseline classification algorithms can successfully predict clicks on the test set.

\subsection{Fairness-aware datasets}

The shortcomings of existing fairness-aware data sets include: age of the data itself, measurement bias, missing values, label leakage, use of data sets for a purpose they were not intended for initially~\citep{le2022survey,hort2023bias}. FairJob dataset is collected in 2024, does not have missing values and represents the real-world application for Fair AI methods.  

From benchmark studies on encoding numerical and categorical features~\citep{Gorishniy2021Revisiting,gorishniy2022embeddings,Grinsztajn2022why,matteucci2023benchmark} and surveys on fairness-aware tabular datasets ~\citep{le2022survey,hort2023bias}, we extract the most used dataset and compare to FairJob in Table~\ref{table:datasets_comparison}. 

\begin{table}[!ht]
\caption{Comparison of FairJob to other most frequently used tabular datasets and most frequently used tabular fairness-aware datasets.}
\label{table:datasets_comparison}
\centering
\begin{tabular}{{@{\ } c @{\ }c@{\ }c@{\ }c @{\ }c @{\ }c @{}}}\toprule
\text{name} & \text{\ \# rows} & \text{\ \# num} & \text{\ \# cat} & \text{\ task type}  & \text{protected attribute} \\ \hline
\text{COMPAS} & 7,214 & 14 & 37 & \text{binclass} & sex, race \\ 
\text{Gesture Phase} & 9,873 & 32 & 0 & \text{multiclass} & - \\ 
\text{Churn Modelling} & 10,000 & 10 & 1 & \text{binclass} & - \\ 
\text{California Housing} & 20,640 & 8 & 0 & \text{regression} & income \\ 
\text{House 16H} & 22,784 & 16 & 0 & \text{regression} & -  \\
\text{Adult} & 48,842 & 6 & 8 & \text{binclass} & sex, race, gender \\ 
\text{Otto Group Products} & 61,878 & 93 & 0 & \text{multiclass} & - \\
\text{Higgs Small} & 98,049 & 28 & 0 & \text{binclass} & - \\
\text{Diabetes} & 101,766 & 10 & 40 & \text{binclass} & gender \\
\text{Facebook Comments Volume} & 197,080 & 50 & 1 & \text{regression} & - \\ 
\text{Santander Customer Transactions} & 200,000 & 200 & 0 & \text{binclass} & - \\
\text{KDD Cencus-Income} & 299,285 & 34 & 7 & \text{binclass} & sex, race \\
\text{Covertype} & 581,012 & 54 & 0 & \text{multiclass} & - \\ 
% \text{Agrawal1} & 1000000 & 9 & 3  & \text{binclass} & + \\
\rowcolor{black!20}[0pt][0pt] \text{FairJob} & 1,072,226  & 36 & 18 & \text{binclass} & gender \\
\text{MSLR-WEB10K (Fold 1)} & 1,200,192 & 136 & 0 & \text{regression} & - \\ 
\bottomrule
\end{tabular}
\end{table}

\section{Broader impact}
\label{section:broader_impact}

Any machine learning system that learns from data runs the risk of introducing unfairness in decision
making. Recent research~\citep{speicher2018potential,lambrecht2019algorithmic,andreou2019measuring,ali2019discrimination} has identified fairness concerns in several AI systems,
especially toward protected groups that are under-represented in the data. Thus, alongside the technical advancements in improving AI systems it crucial that we also focus on ensuring that they work for everyone.

Click prediction is a fundamental task in online advertising and recommendation systems, as noted in recent benchmarking papers~\citep{zhu2021open,mao2023finalmlp}. This is an active research field where improvements in click prediction algorithms lead to improvements in recommendation systems in general~\citep{mao2023finalmlp}. However, improvements in general models do not always translate to specialized models, thus, there is a need to verify algorithms directly with available resources. The FairJob dataset represents a particular case of the click prediction problem, focusing on job-seeking user profiles and job advertising product features. However, many features (user features and categorical product features) are not campaign-specific and can be found in other applications outside job advertising.
By open-sourcing the FairJob dataset, we provide access to a real-world problem of finding a trade-off between utility and fairness in job offer advertising.  We argue that advances in fair machine learning methods should be validated in real-world scenarios, as the distribution of features and their impact on outputs and fairness can vary significantly across different applications. 

As described in Section~\ref{sec:fairness_in_advertising}, advertising, like almost any complex problem, is subject to various biases~\citep{hort2023bias}. We have clearly stated the potential bias sources, analyzed their impact, and discussed possible corrections. While we cannot definitively prove that the properties of the FairJob dataset can be extended to every click prediction problem, it serves as a representation of a specific real-world scenario where the fairness-utility trade-off is encountered. If the FairJob dataset helps to find better methods for improving fairness in job advertising, these methods might also be applicable to other advertising campaigns, as similar user signals may be present in the features used to predict clicks or to associate with the utility-fairness trade-off.

One of the key practical challenges in addressing unfairness in AI systems is that most methods require access to protected demographic features, placing fairness and privacy in tension. We work toward addressing these important challenges by proposing a new benchmarking dataset to improve worst-case performance of protected groups, in the absence of protected group information
but with a proxy variable. 

One limitation of methods in this space is the difficulty of evaluating
their effectiveness when we do not have demographics in a real application. Therefore, while we think developing better debiasing methods is crucial, there remains further challenges in evaluating them.

\end{document}